\documentclass[journal]{IEEEtran}
\usepackage[pdftex]{graphicx}
\usepackage{booktabs}
\usepackage{amsmath}
\usepackage{multirow}

\newcommand{\Tref}[1]{Table~\ref{#1}}

\newcommand{\fref}[1]{Fig.~\ref{#1}}
\newcommand{\Fref}[1]{Figure~\ref{#1}}
\newcommand{\sref}[1]{Sec.~\ref{#1}}
\newcommand{\Sref}[1]{Section~\ref{#1}}
\newcommand{\Aref}[1]{Algorithm~\ref{#1}}
\usepackage{hyperref}
\usepackage{url}
\usepackage{mathrsfs}
\usepackage[utf8]{inputenc} 
\usepackage{booktabs}       
\usepackage{amsfonts}       
\usepackage{nicefrac}       
\usepackage{microtype}      
\usepackage{enumitem}
\usepackage{color}
\usepackage[ruled,vlined]{algorithm2e} 
\usepackage{amssymb}
\usepackage{bm}
\usepackage{xspace}
\usepackage{subfig}
\usepackage[dvipsnames]{xcolor}
\makeatletter
\newcommand{\argmax}{\mathop{\rm arg~max}\limits}
\newcommand{\tblcaption}[1]{\def\@captype{table}\caption{#1}}
\DeclareRobustCommand\onedot{\futurelet\@let@token\@onedot}
\def\@onedot{\ifx\@let@token.\else.\null\fi\xspace}

\def\eg{\emph{e.g}\onedot}

\makeatother

\definecolor{DarkOliveGreen}{rgb}{0.33, 0.42, 0.18}
\newcommand{\codecommentcolor}{DarkOliveGreen} 
\newcommand{\codecomment}[1]{\textcolor{\codecommentcolor}{\texttt{#1}}}

\begin{document}
%
\title{ Unsupervised Learning of Image Segmentation Based on Differentiable Feature Clustering }
%
%
%

\author{Wonjik~Kim$^*$,~\IEEEmembership{Member,~IEEE,}
        Asako~Kanezaki$^*$,~\IEEEmembership{Member,~IEEE,}
        and~Masayuki~Tanaka,~\IEEEmembership{Member,~IEEE}
\thanks{*W. Kim and A. Kanezaki contributed equally to this work.}
\thanks{The authors are with the Tokyo Institute of Technology, Tokyo 152-8550, Japan, Tokyo 135-0061, Japan e-mail: wkim@ok.sc.e.titech.ac.jp; kanezaki@c.titech.ac.jp; mtanaka@sc.e.titech.ac.jp.}
\thanks{The authors are also with the National Institute of Advanced Industrial Science and Technology.}
\thanks{\copyright 20XX IEEE. Personal use of this material is permitted. Permission from IEEE must be obtained for all other uses, in any current or future media, including reprinting/republishing this material for advertising or promotional purposes, creating new collective works, for resale or redistribution to servers or lists, or reuse of any copyrighted component of this work in other works. This paper has been accepted by the IEEE Transactions on Image Processing in July, 2020.}
}

%
%

\markboth{IEEE Transactions on Image Processing, ~Vol.~, No.~, ~}%
{Wonjik \MakeLowercase{\textit{et al.}}: Unsupervised Learning of Image Segmentation}
%



\maketitle

\begin{abstract}
The usage of convolutional neural networks (CNNs) for unsupervised image segmentation was investigated in this study.
Similar to supervised image segmentation, the proposed CNN assigns labels to pixels that denote the cluster to which the pixel belongs.
In unsupervised image segmentation, however, no training images or ground truth labels of pixels are specified beforehand.
Therefore, once a target image is input, the pixel labels and feature representations are jointly optimized, and their parameters are updated by the gradient descent.
In the proposed approach, label prediction and network parameter learning are alternately iterated to meet the following criteria:
(a) pixels of similar features should be assigned the same label,
(b) spatially continuous pixels should be assigned the same label, and 
(c) the number of unique labels should be large.
Although these criteria are incompatible, the proposed approach minimizes the combination of similarity loss and spatial continuity loss to find a plausible solution of label assignment that balances the aforementioned criteria well.
The contributions of this study are four-fold.
First, we propose a novel end-to-end network of unsupervised image segmentation that consists of normalization and an argmax function for differentiable clustering.
Second, we introduce a spatial continuity loss function that mitigates the limitations of fixed segment boundaries possessed by previous work.
Third, we present an extension of the proposed method for segmentation with scribbles as user input, which showed better accuracy than existing methods while maintaining efficiency.
Finally, we introduce another extension of the proposed method: unseen image segmentation by using networks pre-trained with a few reference images without re-training the networks.
The effectiveness of the proposed approach was examined on several benchmark datasets of image segmentation.
\end{abstract}


%
\IEEEpeerreviewmaketitle


\section{Introduction}
\label{sec:introduction}

\IEEEPARstart{I}{mage} segmentation has garnered attention in computer vision research for decades.
The applications of image segmentation include object detection, texture recognition, and image compression.
In supervised image segmentation, a set consisting of pairs of images and pixel-level semantic labels, such as ``sky'' or ``bicycle'', is used for training. 
The objective is to train a system that classifies the labels of the \textit{known} categories for the image pixels.
In contrast, unsupervised image segmentation is used to predict more general labels, such as ``foreground'' and ``background''.
The latter is more challenging than the former. Furthermore, it is extremely difficult to segment an image into an arbitrary number ($\geq$ 2) of plausible regions.
This study considers a problem in which an image is partitioned into an arbitrary number of salient or meaningful regions without any previous knowledge.

Once the pixel-level feature representation is obtained, image segments can be obtained by clustering the feature vectors. 
However, the design of feature representation remains a challenge. 
The desired feature representation depends considerably on the content of the target image. For instance, if the objective is to detect zebras as a foreground, the feature representation should be reactive to black-white vertical stripes.
Therefore, the pixel-level features should be descriptive of the colors and textures of a local region surrounding each pixel.
Recently, convolutional neural networks (CNNs) have been successfully applied to semantic image segmentation in supervised learning scenarios such as autonomous driving and augmented reality games.
CNNs are not often used in completely unsupervised scenarios;
however, they have great potential for extracting detailed features from image pixels, which is necessary for unsupervised image segmentation.
Driven by the high feature descriptiveness of CNNs, a joint learning approach is presented that predicts, for an arbitrary image input, \textit{unknown} cluster labels and learns the optimal CNN parameters for the image pixel clustering.
Subsequently, a group of image pixels in each cluster as a segment is extracted.

The characteristics of the cluster labels that are necessary for good image segmentation are discussed further.
Similar to previous studies on unsupervised image segmentation \cite{unnikrishnan2007toward,yang2008unsupervised},
it is assumed that a good image segmentation solution matches well with a solution that a human would provide.
When a human is asked to segment an image, they would most likely create segments, each of which corresponds to the whole or a salient part of a single object instance.
An object instance tends to contain large regions of similar colors or texture patterns.
Therefore, grouping spatially continuous pixels that have similar colors or texture patterns into the same cluster is a reasonable strategy for image segmentation.
To separate segments from different object instances, it is better to assign different cluster labels to the neighboring pixels of dissimilar patterns.
To facilitate the cluster separation, a strategy in which a large number of unique cluster labels is desired is considered as well.
In conclusion, the following three criteria for the prediction of cluster labels are introduced:
\begin{enumerate}[label=(\alph*)]
\item Pixels of similar features should be assigned the same label.
\item Spatially continuous pixels should be assigned the same label.
\item The number of unique cluster labels should be large.
\end{enumerate}

In this paper, we propose a CNN-based algorithm that jointly optimizes feature extraction functions and clustering functions to satisfy these criteria.
Here, in order to enable end-to-end learning of a CNN, an iterative approach to predict cluster labels using differentiable functions is proposed. The code is available online~\footnote{\url{https://github.com/kanezaki/pytorch-unsupervised-segmentation-tip/}}.

This study is an extension of the previous research published in the international conference on acoustics, speech and signal processing (ICASSP) 2018 \cite{kanezaki2018unsupervised}.
In the previous work, superpixel extraction using simple linear iterative clustering \cite{achanta2012slic} was employed for criterion (b).
However, the previous algorithm had a limitation that the boundaries of the segments were fixed in the superpixel extraction process.
In this study, a spatial continuity loss is proposed as an alternative to mitigate the aforementioned limitation.
In addition, two new applications based on our improved unsupervised segmentation method are introduced: segmentation with user input and utilization of network weights obtained using unsupervised learning of different images.
As the proposed method is completely unsupervised, it segments images based on their nature, which is not always related with the user's intention.
As an exemplar application of the proposed method, scribbles were used as user input and the effect was compared with other existing methods.
Subsequently, the proposed method incurred a high calculation cost to iteratively obtain the segmentation results of a single input image.
Therefore, as another potential application of the proposed method, the network weights pre-trained with several reference images were used.
Once the network weights are obtained from several images using the proposed algorithm, a new unseen image can be segmented by the fixed network, provided it is somewhat similar to the reference images.
The utilization of this technique for a video segmentation task was demonstrated as well.

The contributions of this paper are summarized as follows.
\begin{itemize}
  \item We proposed a novel end-to-end differentiable network of unsupervised image segmentation.
  \item We introduced a spatial continuity loss function that mitigated the limitations of our previous method \cite{kanezaki2018unsupervised}.
  \item We presented an extension of the proposed method for segmentation with scribbles as user input, which showed better accuracy than existing methods while maintaining efficiency.
  \item We introduced another extension of the proposed method: unseen image segmentation by using networks pre-trained with a few reference images without re-training the networks.
\end{itemize}

\section{Related Work}
Image segmentation is the process of assigning labels to all the pixels within an image such that the pixels sharing certain characteristics are assigned the same labels. 
Classical image segmentation can be performed by \eg $k$-means clustering \cite{macqueen1967some}, which is a de facto standard method for vector quantization.
The $k$-means clustering aims to assign the target data to $k$ clusters in which each datum belongs to the cluster with the nearest mean.
The graph-based segmentation method (GS) \cite{felzenszwalb2004efficient} is another example that makes simple greedy decisions of image segmentation.
It produces segmentation results that follow the global features of not being too coarse or too fine based on a particular region comparison function.
Similar to the classical methods, the proposed method in this study aims to perform unsupervised image segmentation. 
In recent, there have been proposed a few methods on learning based unsupervised image segmentation~\cite{liu2014mslrr,xia2017w,croitoru2019unsupervised}.
MsLRR~\cite{liu2014mslrr} is an efficient and versatile approach that can be switched to both unsupervised and supervised methods. MsLRR~\cite{liu2014mslrr} employed superpixels (as our previous work~\cite{kanezaki2018unsupervised}), which caused a limitation that the boundaries are fixed to those of the superpixels.
W-Net~\cite{xia2017w} performs unsupervised segmentation by estimating segmentation from an input image and restoring the input image from the estimated segmentation. Therefore, similar pixels are assigned to the same label, though it does not estimate the boundary of each segment.
Croitoru et al.~\cite{croitoru2019unsupervised} proposed an unsupervised segmentation method based on deep neural network techniques. 
Whereas this method performs binary foreground/background segmentation, our method generates arbitrary number of segments.
A comprehensive survey about deep learning techniques for image segmentation is presented in \cite{ghosh2019understanding}.

The remainder of this section introduces image segmentation with user input, weakly-supervised image segmentation based on CNN, and methods for unsupervised deep learning.

\noindent{\bf Image segmentation with user input:}
\label{sec:related_work_input}
Graph cut is a common method for image segmentation that works by minimizing the cost of a graph where image pixels correspond to the nodes.
This algorithm can be applied to image segmentation with certain user inputs such as scribbles~\cite{boykov2001interactive} and bounding boxes~\cite{lempitsky2009image}.
Image matting is commonly used for image segmentation with user input~\cite{levin2007closed,levin2008spectral} as well.
The distinguishing characteristic of image matting is the soft assignment of pixel labels, whereas, graph cuts produce hard segmentation where every pixel belongs to either the foreground or background.
Constrained random walks~\cite{yang2010user} is proposed to achieve interactive image segmentation with a more flexible user input, which allows scribbles to specify the boundary regions as well as the foreground/background seeds.
Recently, a quadratic optimization problem related to dominant-set clusters has been solved with several types of user input: scribbles, sloppy contours, and bounding boxes \cite{zemene2016interactive}.

The abovementioned methods chiefly produce a binary map that separates image pixels into foreground and background.
In order to apply the graph cut to multi-label segmentation problems, 
the $\alpha$-$\beta$ swap and $\alpha$-expansion algorithms were proposed in \cite{boykov1999fast}.
Both algorithms process repeatedly to find the global minimum of a binary labeling problem.
In $\alpha$-expansion algorithm, an expansion move is defined for a label $\alpha$ to increase the set of pixels that are given this label.
This algorithm finds a local minimum such that no expansion move for any label $\alpha$ yields a labeling with lower energy.
A swap move takes some subset of the pixels presently labeled $\alpha$ and assigns them the label $\beta$ and vice versa for a pair of labels $\alpha$, $\beta$.
The $\alpha$-$\beta$ swap algorithm finds a minimum state such that there is no swap move for any pair of labels $\alpha$, $\beta$ that produces a lower energy labeling.

\noindent{\bf Weakly-supervised image segmentation based on CNN:}
\label{sec:related_weakly}
Semantic image segmentation based on CNNs have been gaining importance in the literature \cite{badrinarayanan2015segnet,chen14semantic,long2015fully,zheng2015conditional}.
As pixel-level annotations for image segmentation are difficult to obtain, weakly supervised learning approaches using object detectors \cite{tighe2013finding,hariharan2014simultaneous,dai2016instance}, object bounding boxes \cite{zhu2014learning,chang2014multiple}, image-level class labels \cite{pathak2015constrained,pourian2015weakly,shi2016weakly,shimoda2016distinct}, or scribbles \cite{lin2016scribblesup, tang2018normalized, tang2018regularized} for training are widely used.

Most of the weakly supervised segmentation algorithms \cite{lin2016scribblesup,zhu2014learning,chang2014multiple,shimoda2016distinct} generate a \textit{training target} from the weak labels and update their models using the generated training set.
Therefore, these methods follow an iterative process that alternates between two steps: (1) gradient descent for training a CNN-based model from the generated target and (2) training target generation by the weak labels.
For example, ScribbleSup \cite{lin2016scribblesup} propagates the semantic labels of scribbles to other pixels using super-pixels so as to completely annotate the images (step 1) and learns a convolutional neural network for semantic segmentation with the annotated images (step 2).
In the case of e-SVM \cite{zhu2014learning}, the segment proposals from bounding box annotations or pixel level annotations using CPMC segments \cite{carreira2011cpmc} (step 1) are generated and the model is trained with the generated segment proposals (step 2).
Shimoda et al. \cite{shimoda2016distinct} estimated class saliency maps using image level annotation (step 1) and applied fully-connected CRF \cite{krahenbuhl2011efficient} with the estimated saliency maps as unary potential (step 2).
These iterative processes are exposed to danger that the convergence is not guaranteed.
The error in training target generation with weak labels might reinforce the entire algorithm to update the model in an undesired direction.
Therefore, recent approaches \cite{tang2018regularized, tang2018normalized, huang2018weakly} for avoiding the error in training target generation with weak labels are proposed.
In this study, to deal with the convergence problem, an \textit{end-to-end} differentiable segmentation algorithm based on a CNN is proposed.

\noindent{\bf Unsupervised deep learning:}
Unsupervised deep learning approaches are mainly focused on learning high-level feature representations using generative models \cite{lee2009unsupervised,le2013building,lee2009convolutional}.
The idea behind these studies is closely related to the conjecture in neuroscience that there exist neurons that represent specific semantic concepts.
In contrast, the application of deep learning to image segmentation and importance of high-level features extracted with convolutional layers are investigated in this study.
Deep CNN filters are known to be effective for texture recognition and segmentation~\cite{cimpoi2015deep,hariharan2015hypercolumns}.

Notably, the convolution filters used in the proposed method are {\it trainable} in the standard backpropagation algorithm, although there are no ground truth labels.
The present study is therefore related to the recent research on deep embedded clustering (DEC) \cite{xie2016unsupervised}.
The DEC algorithm iteratively refines clusters by minimizing the KL divergence loss among the soft-assigned data points with an auxiliary target distribution, whereas, the proposed method simply minimizes the softmax loss based on the estimated clusters.
Similar approaches such as maximum margin clustering~\cite{xu2005maximum} and discriminative clustering~\cite{bach2008diffrac,joulin2010discriminative} have been proposed for semi-supervised learning frameworks;
however, the proposed method is focused on completely unsupervised image segmentation.

\section{Method}
\label{sec:method}



\begin{figure*}[t]
  \begin{center}
    \includegraphics[width=\linewidth]{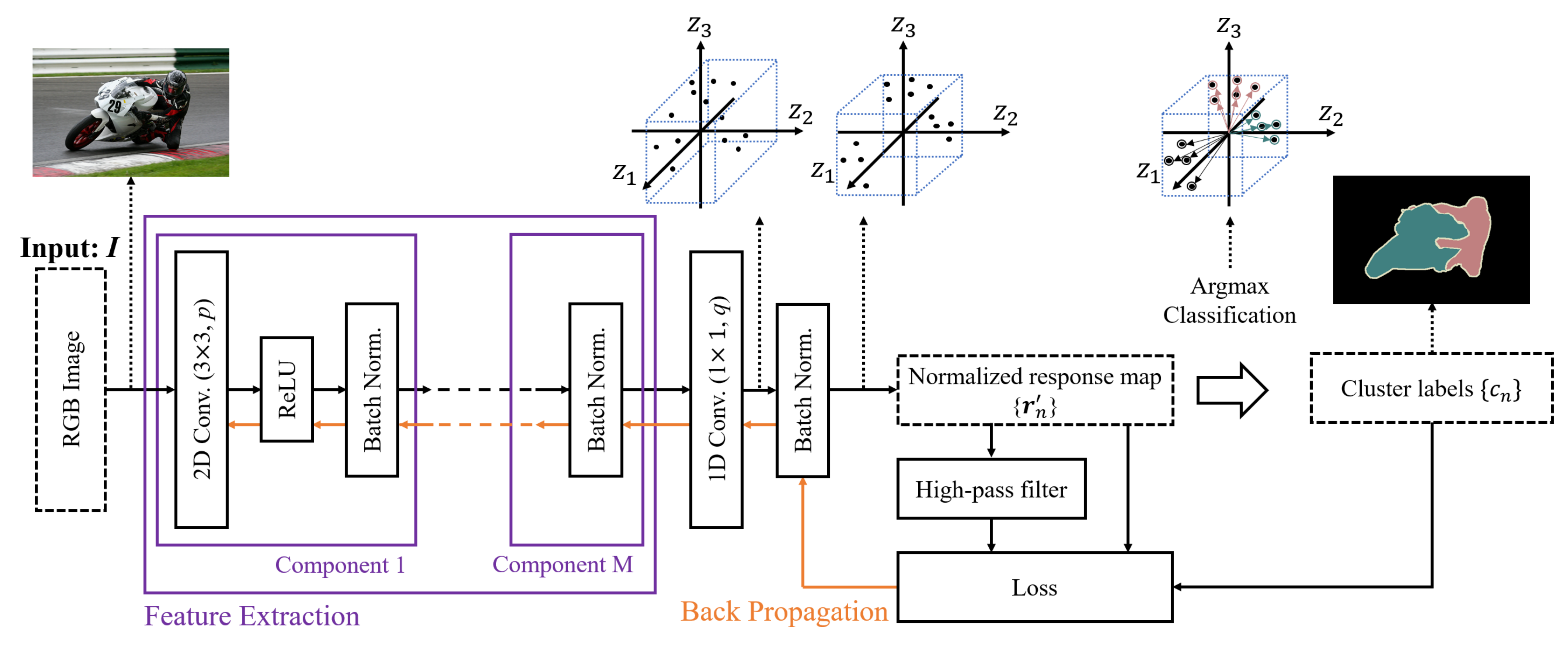}
  \end{center}
  \caption{Illustration of the proposed algorithm for training a CNN.
    Input image $I$ is fed into the CNN to extract deep features $\{ {\bm x}_n \}$ using a feature extraction module.
    Subsequently, one-dimensional (1D) convolutional layer calculates the response vectors $\{ {\bm r}_n \}$ of the features in $q$-dimensional cluster space, where $q=3$ in this illustration.
    Here, $z_1$, $z_2$, and $z_3$ represent the three axes of the cluster space.
    Subsequently, the response vectors are normalized across the axes of the cluster space using a batch normalization function.
    Further, cluster labels $\{ c_n \}$ are determined by assigning the cluster IDs to the response vectors using an argmax function.
    The cluster labels are then used as pseudo targets to compute the feature similarity loss.
    Finally, the spatial continuity loss as well as the feature similarity loss are computed and backpropagated.
  }
  \label{fig:uis}
\end{figure*}

The problem that is solved for image segmentation is described as follows.
For simplicity, let $\{ \}$ denote $\{ \}_{n=1}^N$ unless otherwise noted, where $N$ denotes the number of pixels in an input color image $\mathcal{I}= \{\bm{v}_n \in \mathbb{R}^3 \}$.
Let $f: \mathbb{R}^3 \rightarrow \mathbb{R}^p$ be a feature extraction function and $\{{\bm x}_n \in \mathbb{R}^p\}$ be a set of $p$-dimensional feature vectors of image pixels.
Cluster labels $\{c_n \in \mathbb{Z}\}$ are assigned to all of the pixels by
$
  c_n = g( {\bm x}_n ), 
$
where $g: \mathbb{R}^p \rightarrow \mathbb{Z} $ denotes a mapping function.
Here, $g$ can be an assignment function that returns the label of the cluster centroid closest to ${\bm x}_n$.
For the case in which $f$ and $g$ are fixed,
$\{c_n\}$ are obtained using the abovementioned equation.
Conversely, if $f$ and $g$ are trainable whereas $\{c_n\}$ are specified (fixed), then the aforementioned equation can be regarded as a standard supervised classification problem.
The parameters for $f$ and $g$ in this case can be optimized by gradient descent if $f$ and $g$ are differentiable.
However, in the present study, \textit{unknown} $\{c_n\}$ are predicted while training the parameters of $f$ and $g$ in a completely unsupervised manner.
To put this into practice, the following two sub-problems were solved: 
prediction of the optimal $\{c_n\}$ with fixed $f$ and $g$ and training of the parameters of $f$ and $g$ with fixed $\{c_n\}$.

Notably, the three criteria introduced in \sref{sec:introduction} are incompatible and are never satisfied perfectly. One possible solution for addressing this problem using a classical method is: applying $k$-means clustering to $\{{\bm x}_n\}$ for (a), performing graph cut algorithm~\cite{boykov1999fast} using distances to centroids for (b), and determining $k$ in $k$-means clustering using a non-parametric method for (c).
However, these classical methods are only applicable to \textit{fixed} $\{{\bm x}_n\}$ and therefore the solution can be suboptimal.
Therefore, a CNN-based algorithm is proposed to solve the problem.
The feature extraction functions for $\{{\bm x}_n\}$ and $\{c_n\}$ are jointly optimized in a manner that satisfies all the aforementioned criteria.
In order to enable end-to-end learning of a CNN, an iterative approach to predict $\{c_n\}$ using differentiable functions is proposed.

A CNN structure is proposed, as shown in \fref{fig:uis}, along with a loss function to satisfy the three criteria described in \sref{sec:introduction}.
The concept of the proposed CNN architecture for considering criteria (a) and (c) is detailed in \Sref{sec:architecture}.
The concept of loss function for solving criteria (a) and (b) is presented in \Sref{sec:loss}.
The details of training a CNN using backpropagation are described in \sref{sec:backprop}.


\subsection{Network architecture}
\label{sec:architecture}

\subsubsection{Constraint on feature similarity}
\label{sec:(a)}
We consider the first criterion of assigning the same label to pixels with similar characteristics.
The proposed solution is to apply a linear classifier that classifies the features of each pixel into $q$ classes.
In this study, we assume the input to be an RGB image $\mathcal{I}= \{\bm{v}_n \in \mathbb{R}^3 \}$, where each pixel value is normalized to $[0,1]$.
A $p$-dimensional feature map $\{{\bm x}_n\}$ is computed from $\{{\bm v}_n\}$ through $M$ convolutional components, each of which consists of a two-dimensional (2D) convolution, ReLU activation function, and a batch normalization function, 
where a batch corresponds to $N$ pixels of a single input image.
Here, we set $p$ filters of region size $3 \times 3$ for all of the $M$ components.
Notably, these components for feature extraction can be replaced by alternatives such as fully convolutional networks (FCN)~\cite{long2015fully}.
Subsequently, a response map $\{{\bm r}_n = W_c {\bm x}_n \}$ is obtained by applying a linear classifier, where $W_c \in \mathbb{R}^{q \times p}$.
The response map is then normalized to $\{{\bm r}'_n\}$ such that $\{{\bm r}'_n\}$ has zero mean and unit variance.  
The motivation behind the normalization process is described in \sref{sec:(c)}.
Finally, the cluster label $c_n$ for each pixel is obtained by selecting the dimension that has the maximum value in ${\bm r}'_n$.
This classification rule is referred to as the argmax classification.
This processing corresponds intuitively to the clustering of feature vectors into $q$ clusters.
The $i$th cluster of the final responses $\{{\bm r}'_n\}$ can be written as:
\begin{equation}
  C_i = \{ {\bm r}'_n \in \mathbb{R}^q ~ | ~~ r'_{n,i} \geq r'_{n,j}, ~~ \forall j \},
  \nonumber
\end{equation}
where $r'_{n,i}$ denotes the $i$th element of ${\bm r}'_n$.
This is equivalent to assigning each pixel to the closest point among the $q$ representative points, which are placed at infinite distance on the respective axis in the $q$-dimensional space.
Notably, $C_i$ can be $\emptyset$, and therefore the number of unique cluster labels can arbitrarily range from 1 to $q$. 


\subsubsection{Constraint on the number of unique cluster labels}
\label{sec:(c)}

In unsupervised image segmentation, there is no clue as to how many segments should be generated in an image.
Therefore, the number of unique cluster labels should be adaptive to the image content.
As described in \sref{sec:(a)}, the proposed strategy is to classify pixels into an arbitrary number $q'$ $(1 \leq q' \leq q)$ of clusters,
where $q$ is the maximum possible value of $q'$.
A large $q'$ indicates oversegmentation, whereas a small $q'$ indicates undersegmentation.
To train a neural network, we set a large number to the initial (maximum) number of cluster labels $q$.
Then, in the iterative update process, similar or spatially close pixels are integrated by considering feature similarity and spatial continuity constraints.
This phenomenon leads to reduce the number of unique cluster labels $q'$, 
even though there is no explicit constraint on $q'$.

As shown in \fref{fig:uis}, the proposed clustering function based on argmax classification corresponds to $q'$-class clustering, where $q'$ anchors correspond to a subset of $q$ points at infinity on the $q$ axes.
The aforementioned criteria (a) and (b) only facilitate the grouping of pixels, which could lead to a simple solution that $q'=1$.
To prevent this kind of undersegmentation failure, a third criterion (c) is introduced, which is the preference for a large $q'$.
The proposed solution is to insert the \textit{intra-axis} normalization process for the response map $\{{\bm r}_n\}$ before assigning cluster labels using the argmax classification.
Here, batch normalization~\cite{ioffe2015batch} is used.
This operation, also known as whitening, converts the original responses $\{{\bm r}_n\}$ to $\{{\bm r}'_n\}$, where each axis has zero mean and unit variance.
This gives each $r'_{n,i}$ $(i=1,\dots,q)$ an even chance to be the maximum value of ${\bm r}'_n$ across the axes.
Although this operation does not guarantee that every cluster index $i$ $(i=1,\dots,q)$ achieves the maximum value for any $n$ $(n=1, \dots, N)$, nevertheless, because of this operation, many cluster indices will achieve the maximum value for any $n$ $(n=1, \dots, N)$.
Consequently, this intra-axis normalization process gives the proposed system a preference for a large $q'$.


\subsection{Loss function}
\label{sec:loss}
The proposed loss function $L$ consists of a constraint on feature similarity and a constraint on spatial continuity, denoted as follows:
\begin{equation}
\begin{aligned}
\label{eq:loss}
    L \ \: =  \ \: {\underbrace{ L_{\rm{sim}}( \{ {\bm r}'_n, c_n \} ) }_{\text{\makebox[0pt]{feature similarity}}}} \ \: + \ \: \mu{\underbrace{ L_{\rm{con}}(\{ {\bm r}'_n \}) }_{\text{\makebox[0pt]{spatial continuity}}}},
\end{aligned}
\end{equation}
where $\mu$ represents the weight for balancing the two constraints.
%
Although the proposed method is a completely unsupervised learning method, the use of the proposed method with scribbles as user input was also investigated.
In the case with segmentation using scribble information, the loss function \eqref{eq:loss} is simply modified using another weight $\nu$ as follows:
\begin{equation}
\begin{aligned}
\label{eq:entireloss}
    L =  {\underbrace{ L_{\rm{sim}}( \{ {\bm r}'_n, c_n \} ) }_{\text{\makebox[0pt]{feature similarity}}}} + \mu{\underbrace{ L_{\rm{con}}( \{ {\bm r}'_n \} ) }_{\text{\makebox[0pt]{spatial continuity}}}} + \nu{\underbrace{ L_{\rm{scr}}( \{ {\bm r}'_n, s_n, u_n \} ) }_{\text{\makebox[0pt]{scribble information}}}}.
\end{aligned}
\end{equation}
Each component of the abovementioned function is described in their respective sections below.


\subsubsection{Constraint on feature similarity}
\label{sec:featuresim}

As described in \sref{sec:(a)}, the cluster labels $\{c_n\}$ are obtained by applying the argmax function to the normalized response map $\{{\bm r}'_n\}$.
The cluster labels are further utilized as the pseudo targets.
In the proposed approach, the following cross entropy loss between $\{{\bm r}'_n\}$ and $\{c_n\}$ is calculated as the constraint on feature similarity:
\begin{equation}
  L_{\rm{sim}}(\{ {\bm r}'_n, c_n \}) = \sum_{n=1}^{N} \sum_{i=1}^{q} - \delta(i - c_n) \ln {r'_{n,i}},
  \nonumber
\end{equation}
where
\begin{equation}
  \; \delta(t) = \begin{cases} 1 & \textrm{if} \; t = 0 \\
                                 0 & \textrm{otherwise}.
  \end{cases}
  \nonumber
\end{equation}
The objective behind this loss function is to \textit{enhance} the similarity of the similar features.
Once the image pixels are clustered based on their features, the feature vectors within the same cluster should be similar to each other, and the feature vectors from different clusters should be different from each other.
Through the minimization of this loss function, the network weights are updated to facilitate extraction of more efficient features for clustering.


\subsubsection{Constraint on spatial continuity}
\label{sec:continuity}

The elementary concept of image pixel clustering is to group similar pixels into clusters, as shown in \sref{sec:(a)}.
In image segmentation, however, it is preferable for the clusters of image pixels to be spatially continuous.
An additional constraint is introduced that favors the cluster labels to be the same as those of the neighboring pixels.

In a similar manner to \cite{shibata2017misalignment}, we consider the L1-norm of horizontal and vertical differences of the response map $ \{{\bm r}'_n\}$ as a spatial constraint.
We can implement the process by a differential operator.
More specifically, the spatial continuity loss $L_{\rm{con}}$ is defined as follows:
\begin{equation}
  L_{\rm{con}}( \{ {\bm r}'_n \} ) = \sum_{\xi=1}^{W-1} \sum_{\eta=1}^{H-1} \! \mid \mid {\bm r}'_{\xi+1,\eta} - {\bm r}'_{\xi,\eta} \mid \mid _1 \! + \! \mid \mid {\bm r}'_{\xi,\eta+1} - {\bm r}'_{\xi,\eta} \mid \mid _1,
  \nonumber
  \label{eq:L_con}
\end{equation}
where $W$ and $H$ represent the width and height of an input image, whereas ${\bm r}'_{\xi,\eta}$ represents the pixel value at $(\xi,\eta)$ in the response map $\{{\bm r}'_n\}$.

By applying the spatial continuity loss $L_{\rm{con}}$, an excessive number of labels due to complicated patterns or textures can be suppressed.


\subsubsection{Constraint on scribbles as user input}
\label{sec:scribble}

Image segmentation technique with scribble information has been researched extensively \cite{yang2010user, lin2016scribblesup, tang2018normalized, tang2018regularized}.
In the proposed approach, scribble loss $L_{\rm{scr}}$ as partial cross entropy was introduced as follows: 
\begin{equation}
\label{eq:scrloss}
  L_{\rm{scr}}( \{ {\bm r}'_n, s_n, u_n \} ) = \sum_{n=1}^{N} \sum_{i=1}^{q} - u_n \delta(i - s_n) \ln {r'_{n,i}},
  \nonumber
\end{equation}
where $u_n = 1$ if the $n$th pixel is a scribbled pixel, otherwise it is $0$, and $s_n$ denotes the scribble label for each pixel.


\begin{algorithm}[t]
{\small
  
  \KwIn{ \ $\mathcal{I}=\{ \bm{v}_n \in \mathbb{R}^3 \}$ \ \ \ \ \ \ \ \ \ \ \ \ \ \ \codecomment{// RGB image} \\
  	\hspace{11mm} $\mu$ \ \ \ \ \ \ \ \ \ \ \ \ \ \ \ \ \ \ \ \ \ \ \ \ \ \ \ \ \ \ \ \codecomment{// weight for $L_{\rm{con}}$} \\
  	} 
  \KwOut{ \ $\mathcal{L}=\{c_n \in \mathbb{Z}\}$ \ \ \ \ \ \ \ \ \ \ \ \ \ \codecomment{// Label image}} 
  
  $\{ W_m, \bm{b}_m \}_{m=1}^M \ \gets$ \texttt{ Init() } \ \ \ \ \ \ \codecomment{// Initialize} \\
  $ W_c  \ \gets$ \texttt{ Init() } \ \ \ \ \ \ \ \ \ \ \ \ \ \ \ \ \ \ \ \ \codecomment{// Initialize} \\
  
  \For{\ $t=1$ \ \KwTo $T$}{
    
\hspace{-2mm}    $\{ \bm{x}_n \} \gets$ \texttt{GetFeats( $\{ \bm{v}_n \}, \{ W_m, \bm{b}_m \}_{m=1}^M $ )} \\
\hspace{-2mm}    $\{ \bm{r}_n \} \gets \{ \ W_c \bm{x}_n \ \}$ \\
\hspace{-2mm}    $\{ \bm{r}'_n \} \gets $ \texttt{Norm(  $\{ \bm{r}_n \}$ )} \ \ \ \ \ \ \ \ \ \codecomment{// Batch norm.} \\
\hspace{-2mm}    $\{ c_n \} \gets \{ \ \argmax_{i}{ \ r'_{n,i} } \ \}$ \ \ \ \ \ \ \ \ \codecomment{//\hspace{1mm}Assign labels} \\

\hspace{-2mm}    $L \gets L_{\rm{sim}}( \{ \bm{r'}_n, c_n \} ) + \mu L_{\rm{con}}( \{ \bm{r'}_n \} )$ \\
\hspace{-2mm}    $\{ W_m, \bm{b}_m \}_{m=1}^M, W_c \gets$ \texttt{Update( $L$ )} \\

  }
}  
  \caption{Unsupervised image segmentation}
  \label{alg:uis}
\end{algorithm}

\subsection{Learning network by backpropagation}
\label{sec:backprop}
In this section,  the method of training the network for unsupervised image segmentation is described.
Once a target image is input, the following two sub-problems are solved: 
the prediction of cluster labels with fixed network parameters and the training of network parameters with the (fixed) predicted cluster labels.
The former corresponds to the forward process of a network followed by the proposed architecture described in \sref{sec:architecture}.
The latter corresponds to the backward process of a network based on the gradient descent.
Subsequently, we calculate and backpropagate the loss $L$ described in \sref{sec:loss} to update the parameters of the convolutional filters $\{ W_m \}_{m=1}^M$ as well as the parameters of the classifier $W_c$.
In this study, a stochastic gradient descent with momentum is used for updating the parameters.
The parameters are initialized with the Xavier initialization~\cite{glorot2010understanding}, which samples values from the uniform distribution normalized according to the input and output layer size.
This forward-backward process is iterated $T$ times to obtain the final prediction of the cluster labels $\{c_n\}$.
\Aref{alg:uis} shows the pseudocode for the proposed unsupervised image segmentation algorithm.
Since this iterative process requires a little computation time, we further introduce a use of the proposed method with one or several reference images.
Provided that a target image is somewhat similar to the reference images, the fixed network weights trained with those images as pre-processing can be reused.
The effectiveness of the use of reference images is investigated in \sref{sec:fixweight}.

As shown in \fref{fig:uis}, the proposed CNN network is composed of basic functions. 
The most distinctive part of the proposed CNN is the existence of the batch normalization layer between the final convolution layer and the argmax classification layer.
Unlike the supervised learning scenario, in which the target labels are fixed, the batch normalization of responses over axes is necessary to obtain reasonable labels $\{c_n\}$ (see \sref{sec:(c)}).
Furthermore, in contrast to supervised learning, there are multiple solutions of $\{c_n\}$ with different network parameters that achieve near zero loss.
The value of the learning rate takes control over the balance between the parameter updates and clustering, which leads to different solutions of $\{c_n\}$.
We set the learning rate to $0.1$ with a momentum of $0.9$.

\section{Experimental results}

As mentioned in \sref{sec:introduction}, a spatial continuity loss is proposed as described in \sref{sec:continuity} as an alternative to superpixel extraction used in our previous study~\cite{kanezaki2018unsupervised}.
The effectiveness of the continuity loss was evaluated by comparing it with \cite{kanezaki2018unsupervised} as well as other classical methods discussed in \sref{sec:contloss}.
Additionally, the use of the proposed method with scribble input in \sref{sec:weaklyexp} and with reference images in \sref{sec:fixweight} was demonstrated. 
The number of convolutional layers $M$ was set to $3$ and $p = q = 100$ for all of the experiments.
For the loss function, different $\mu$ were set for each experiment: $\mu = 5$ for PASCAL VOC 2012 and BSD500 in \Sref{sec:contloss} and \Sref{sec:fixweight}, $\mu = 50$ for iCoseg and BBC Earth in \Sref{sec:fixweight}, $\mu = 100$ for pixabay in \Sref{sec:fixweight}, and $\mu = 1$ for \Sref{sec:weaklyexp}.
The results of all the experiments were evaluated by the mean intersection over union (mIOU).
Here, mIOU was calculated as the mean IOU of each segment in the ground truth (GT) and the estimated segment that had the largest IOU with the GT segment.
Notably, the object category labels in PASCAL VOC 2012 dataset \cite{everingham2015pascal} were ignored and each segment  along with the background region was treated as an individual segment.

\subsection{Effect of continuity loss}
\label{sec:contloss}

\begin{table*}
 \caption{Comparison of mIOU for unsupervised segmentation on PASCAL VOC 2012 and BSD500. The best scores are shown in bold and the second-best scores are underlined. }
 \begin{center}
 {\normalsize
  \begin{tabular}{lrrrrr}
  \toprule
      Method & VOC 2012 & BSD500 all & BSD500 fine & BSD500 coarse & mean \\ \midrule
    $k$-means clustering, $k=2$ & 0.3166 & 0.1223 & 0.0865 & 0.1972 & 0.1807 \\
    $k$-means clustering, $k=17$ & 0.2383 & 0.2404 & 0.2208 & 0.2648 & 0.2411 \\
    Graph-based Segmentation (GS) \cite{felzenszwalb2004efficient}, $\tau=100$ & 0.2682  & {\bf 0.3135} & {\bf 0.2951} & 0.3255 & 0.3006 \\
    Graph-based Segmentation (GS) \cite{felzenszwalb2004efficient}, $\tau=500$ & {\bf 0.3647}  & 0.2768 & 0.2238 & \underbar{0.3659} & \underbar{0.3078} \\
    IIC \cite{ji2019invariant}, $k=2$ & 0.2729 & 0.0896 & 0.0537 & 0.1733 & 0.1474 \\
    IIC \cite{ji2019invariant}, $k=20$ & 0.2005 & 0.1724 & 0.1513 & 0.2071 & 0.1828 \\
    Ours w/ superpixels \cite{kanezaki2018unsupervised} & 0.3082 & 0.2261 & 0.1690 & 0.3239 & 0.2568 \\
    Ours w/ continuity loss, $\mu=5$ & \underbar{0.3520} & \underbar{0.3050} & \underbar{0.2592} & {\bf 0.3739} & {\bf 0.3225} \\ \bottomrule

     
  \end{tabular}
 }
 \end{center}
 \label{table:closs}
\end{table*}

\begin{table}
 \caption{Parameter search on PASCAL VOC 2012.}
 \begin{center}
 {\small
  \begin{tabular}{ccccccc}
  \toprule
    \multicolumn{7}{c}{unsupervised segmentation} \\ \midrule
    $\mu$ & 0.1 & 0.5 & 1 & 5 & 10 & 20 \\ \midrule
    mIOU & 0.3340 & 0.3433 & 0.3449 & {\bf 0.3520} & 0.3483 & 0.3438 \\ \toprule
    \multicolumn{7}{c}{segmentation with user input} \\ \midrule
    $\nu$ & 0.1 & 0.5 & 1 & 5 & 10 & 20 \\ \midrule
    mIOU & 0.4774 & {\bf 0.6174} & 0.5994 & 0.5298 & 0.4982 & 0.4650 \\ \bottomrule
  \end{tabular}
 }
 \end{center}
 \label{table:prs}
\end{table}

\begin{table}
 \caption{Ablation studies on $L_{\rm{con}}$ and batch normalization.}
 \begin{center}
 {\normalsize
  \begin{tabular}{cccrrrr}
  \toprule
     &  &  & & \multicolumn{3}{c}{BSD500} \\ 
    $L_{\rm{sim}}$ \hspace{-2mm} & $L_{\rm{con}}$ \hspace{-2mm} & BN \hspace{-2mm} & VOC2012 & all & fine & coarse \\ \midrule
    \checkmark & & & 0.3312 & 0.2279 & 0.1928 & 0.2932 \\
    \checkmark & \checkmark & & 0.3340 & 0.2199 & 0.1832 & 0.2931 \\
    \checkmark & & \checkmark & \underbar{0.3358} & \underline{0.3007} & {\bf 0.2619} & \underline{0.3506} \\
    \checkmark & \checkmark & \checkmark & {\bf 0.3520} & {\bf 0.3050} & \underline{0.2592} & {\bf 0.3739} \\ \bottomrule
  \end{tabular}
 }
 \end{center}
 \label{table:abl study}
\end{table}

\begin{figure*}[t]
  \begin{center}
  \subfloat[Input image]{
    \includegraphics[width=22mm]{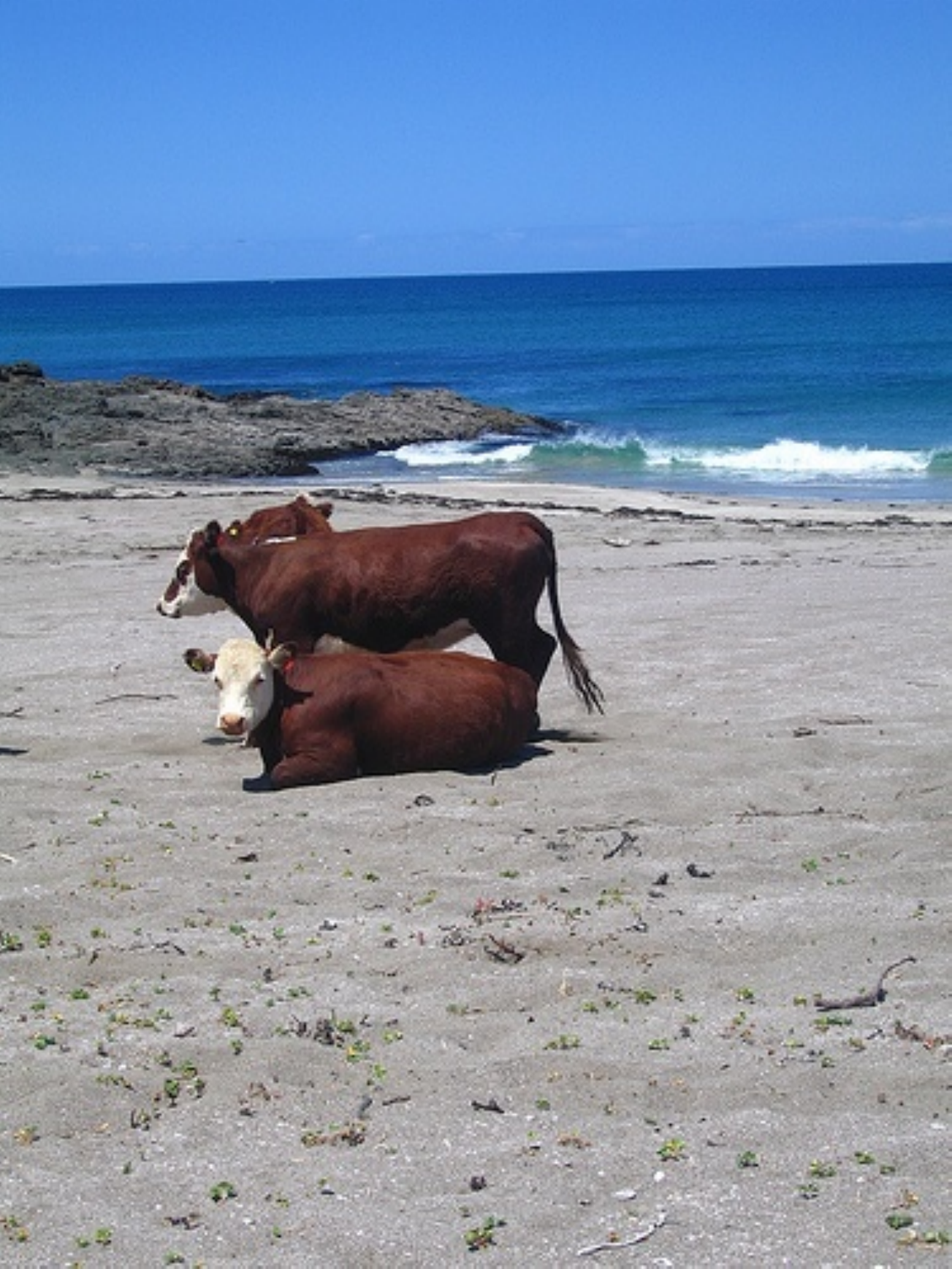}
  }
  \subfloat[Ground truth]{
    \includegraphics[width=22mm]{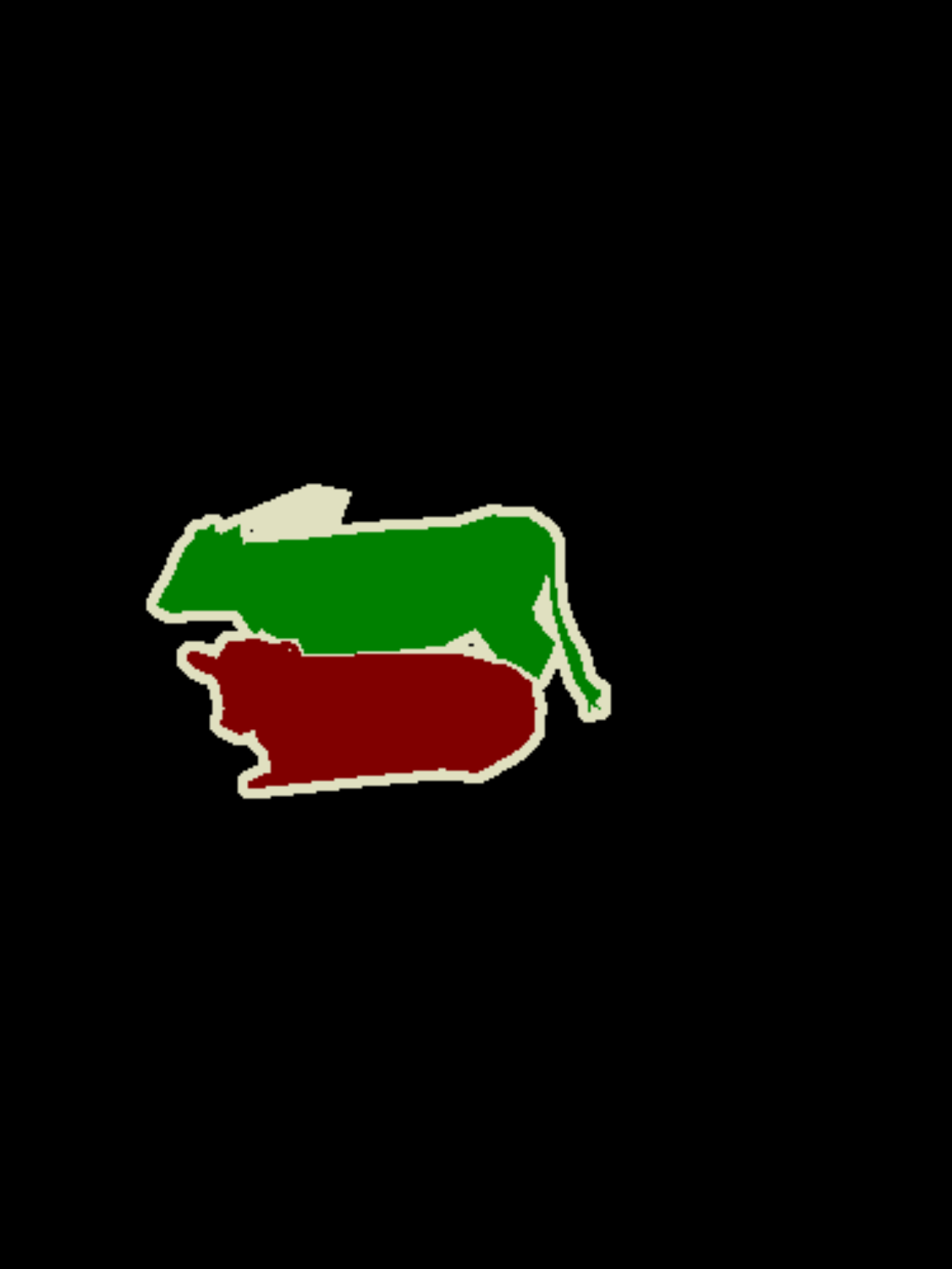}
  }
  \subfloat[$\mu=1$]{
    \includegraphics[width=22mm]{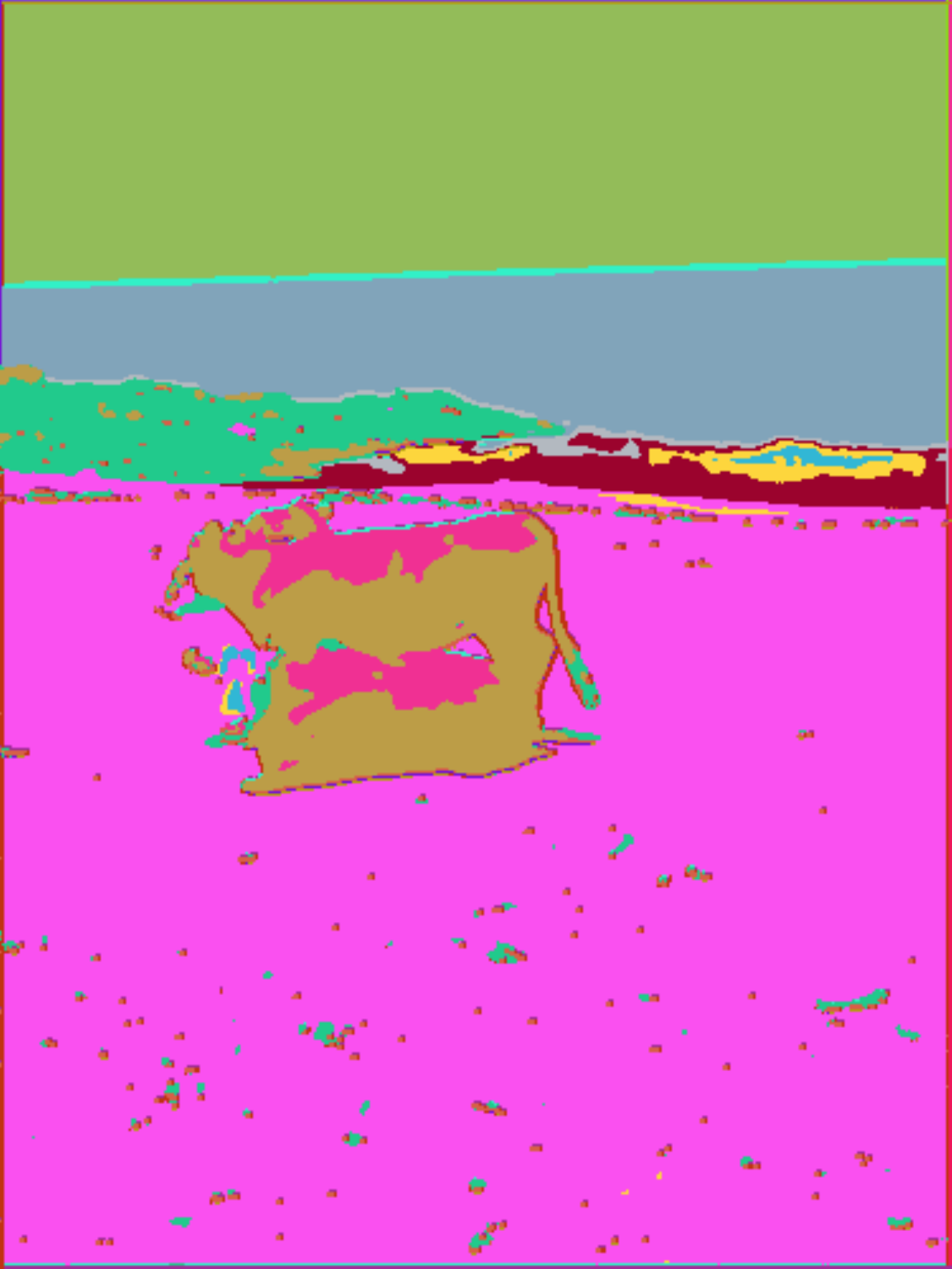}
  }
  \subfloat[$\mu=5$]{
    \includegraphics[width=22mm]{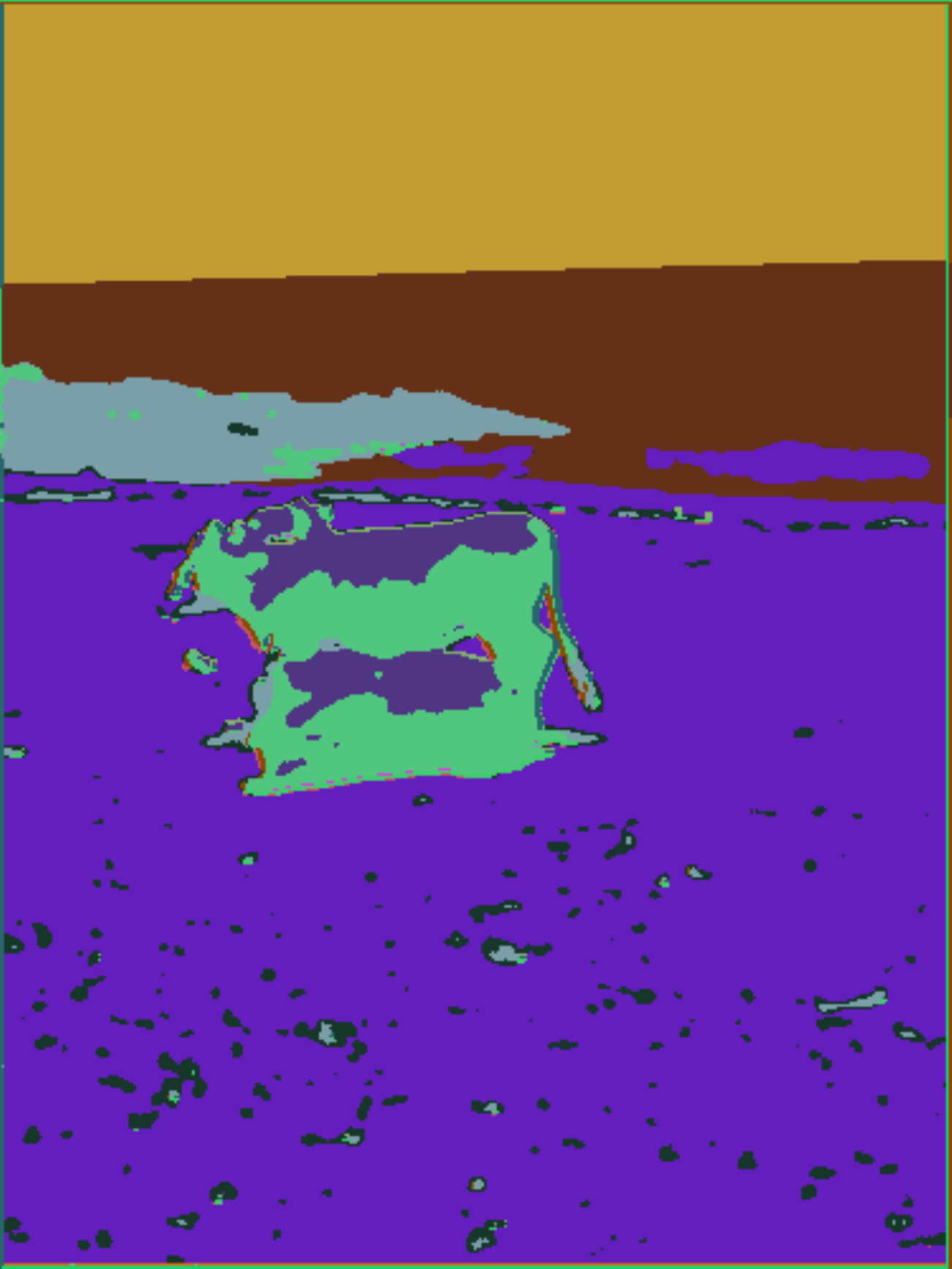}
  }
  \subfloat[$\mu=10$]{
    \includegraphics[width=22mm]{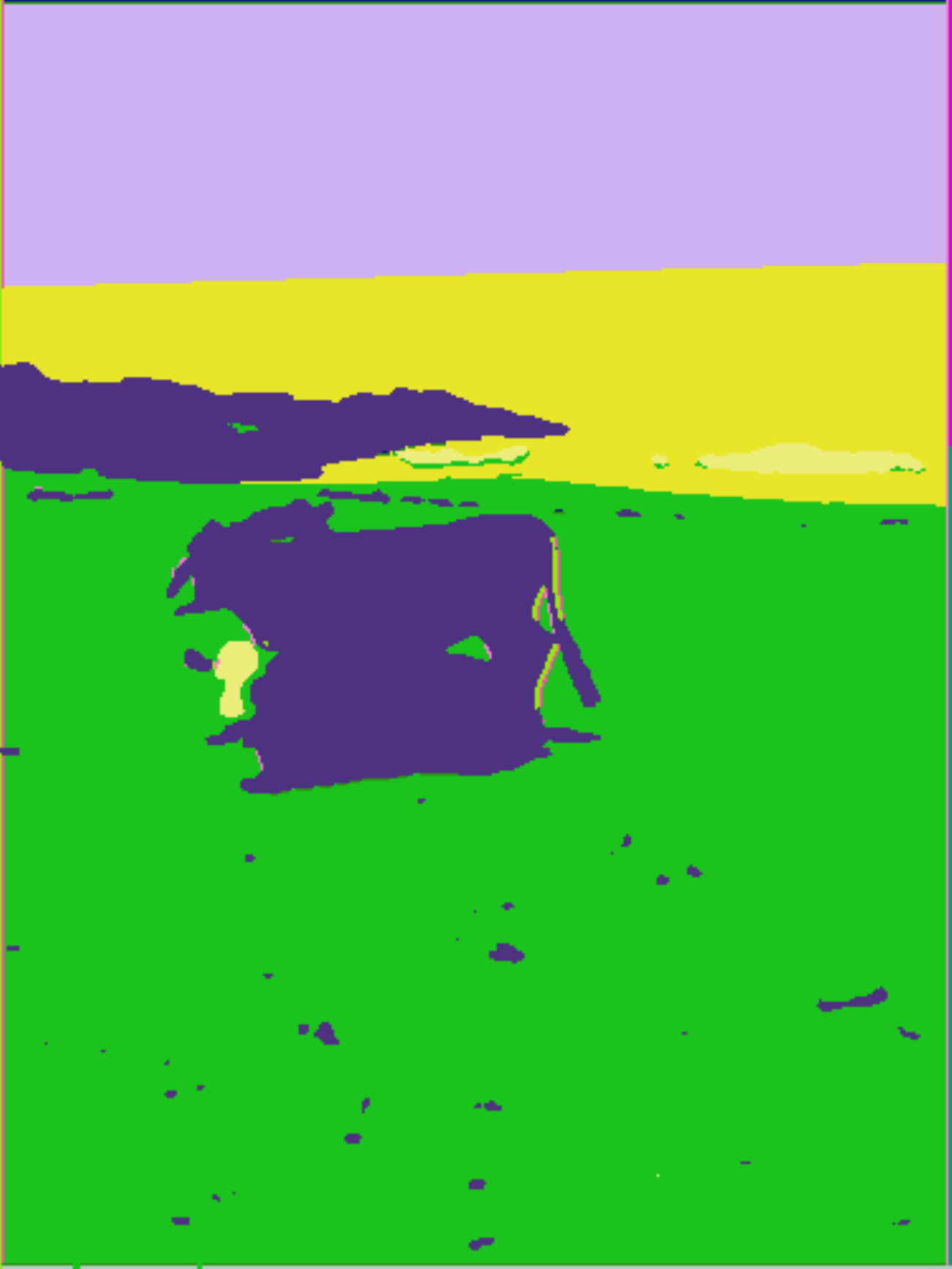}
  }
  \subfloat[$\mu=50$]{
    \includegraphics[width=22mm]{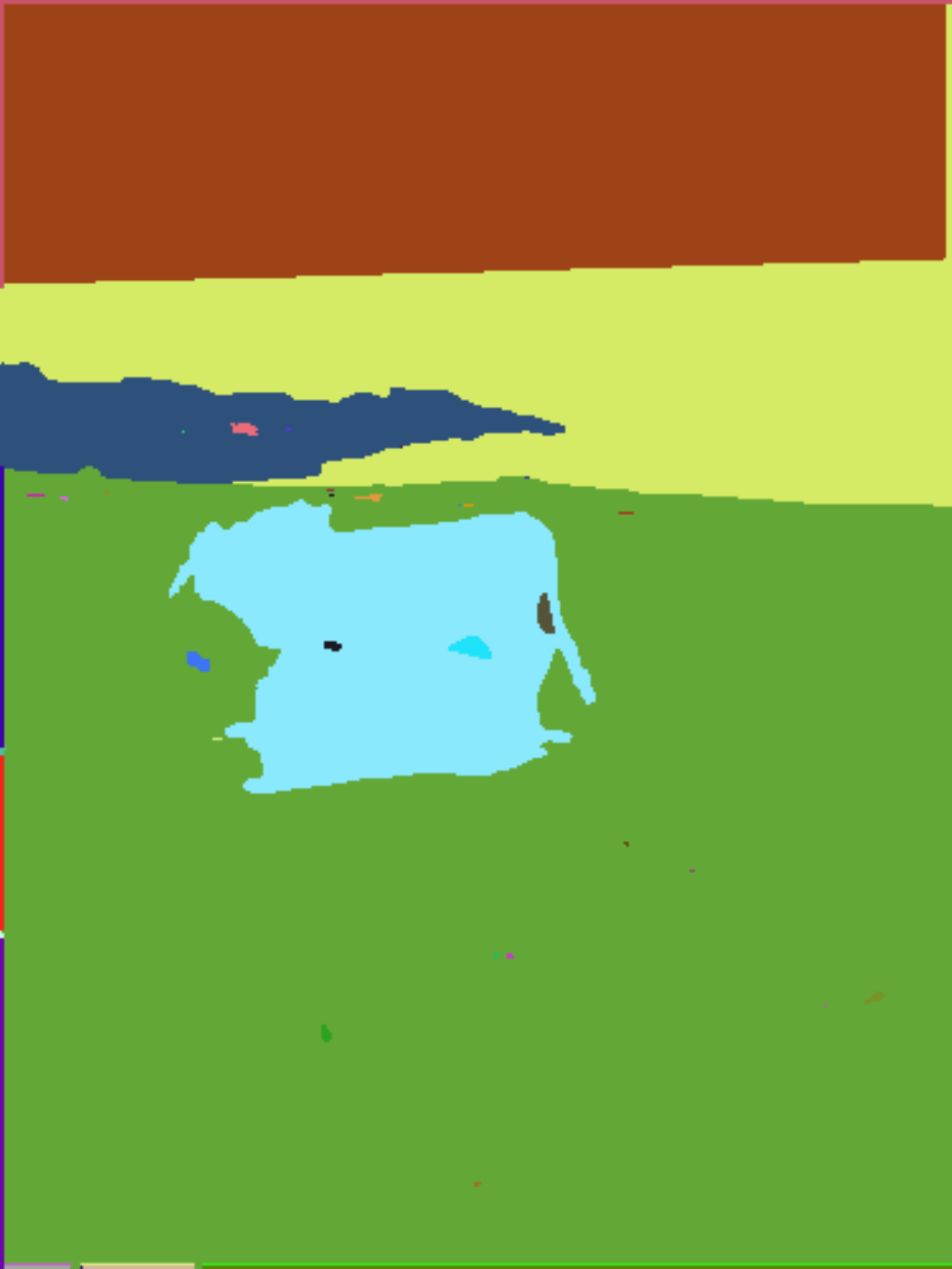}
    \label{fig:comparison_parameter_con_f}
  }
  \subfloat[$\mu=100$]{
    \includegraphics[width=22mm]{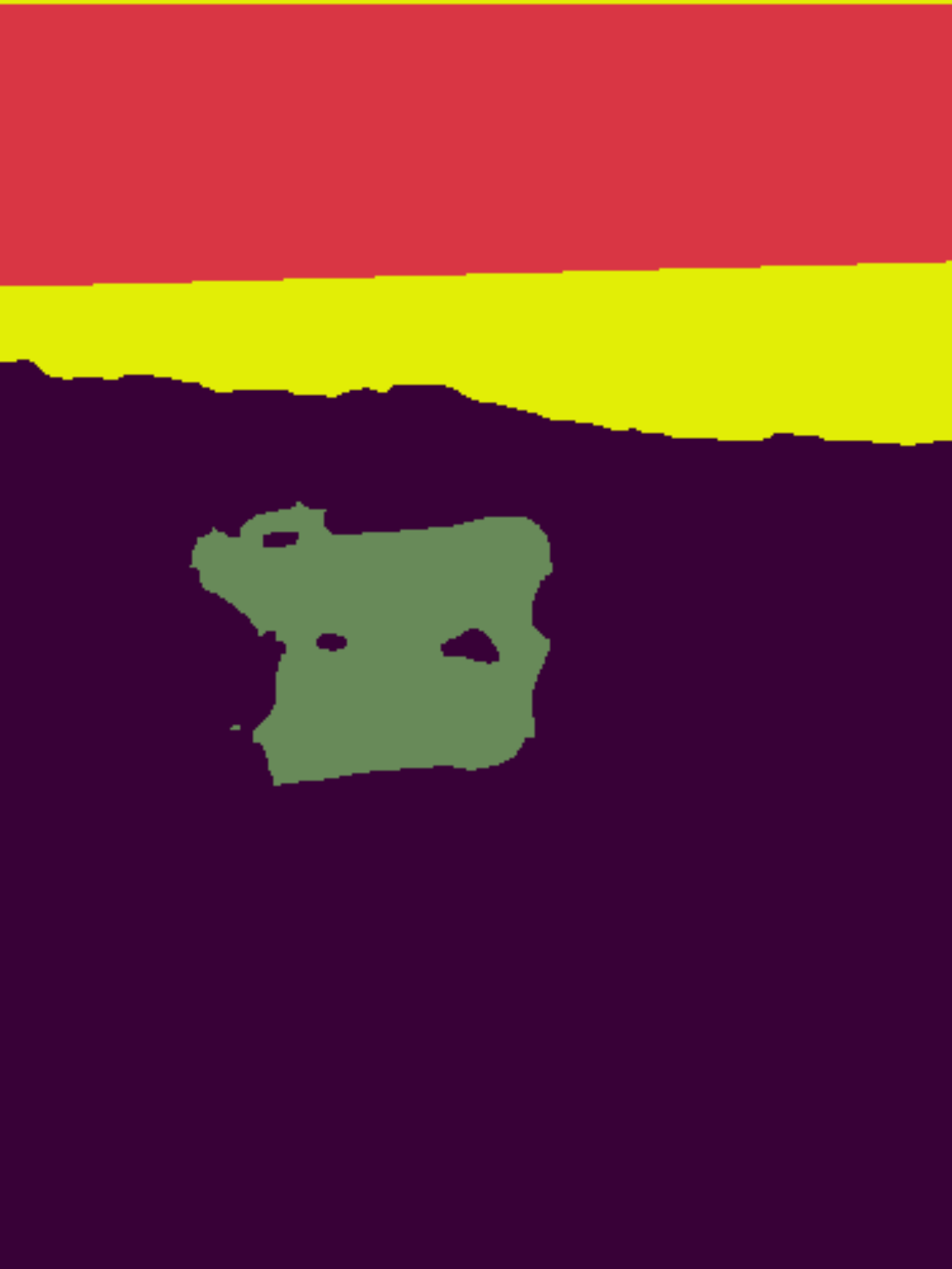}
  }
  \end{center}
  \caption{Effect of continuity loss with different $\mu$ values. Different segments are shown in different colors.}
  \label{fig:comparison_parameter_con}
\end{figure*} 

\begin{figure*}[t]
  \begin{center}
    \includegraphics[width=\linewidth]{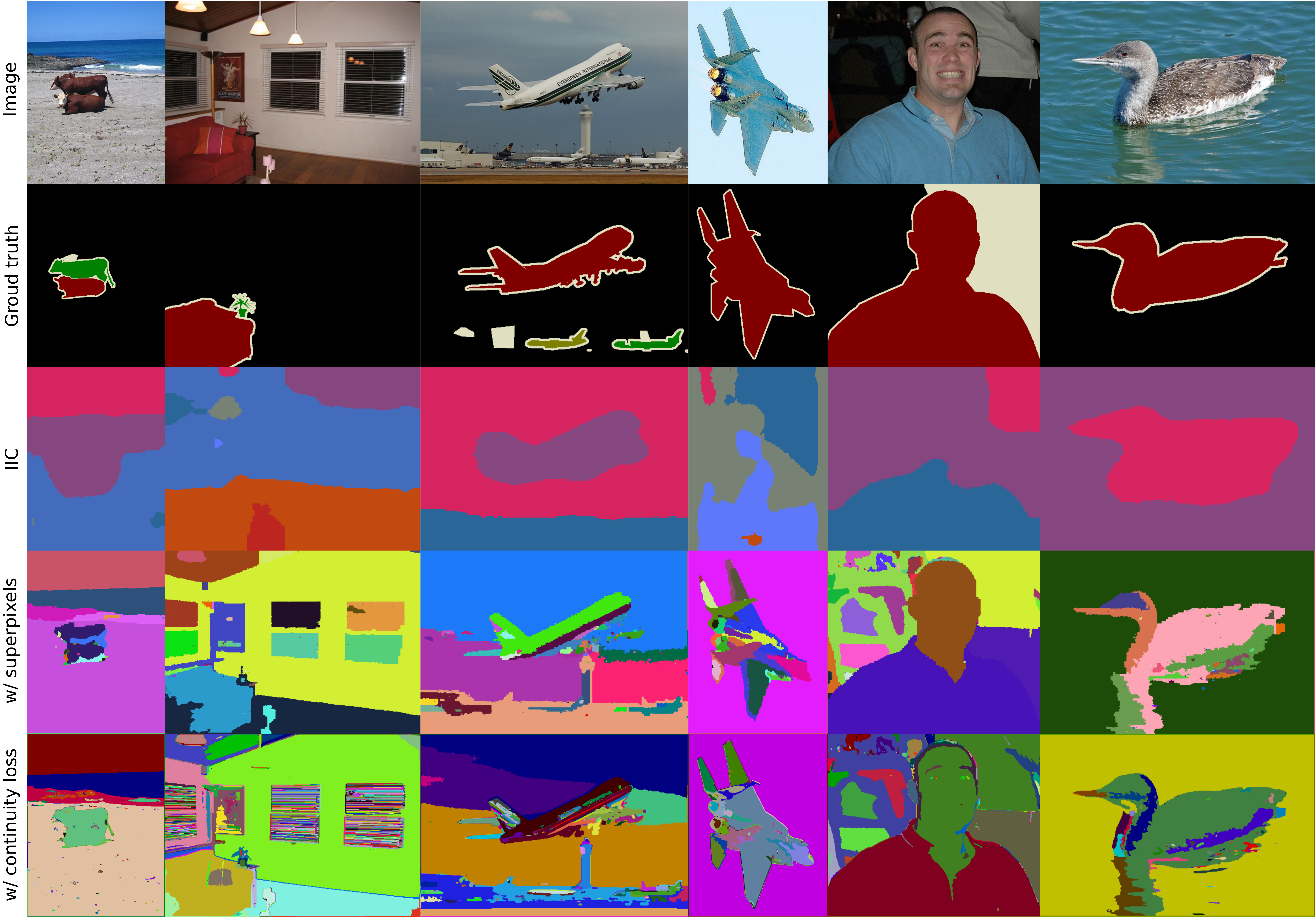}
  \end{center}
  \caption{Comparison of unsupervised segmentation results on PASCAL VOC 2012. The method with superpixels corresponds to the previous method proposed in \cite{kanezaki2018unsupervised}. Different segments are shown in different colors.}
  \label{fig:result_image_con}
\end{figure*} 

\begin{figure*}[t]
  \begin{center}
  \subfloat[Input image]{
    \includegraphics[width=21mm]{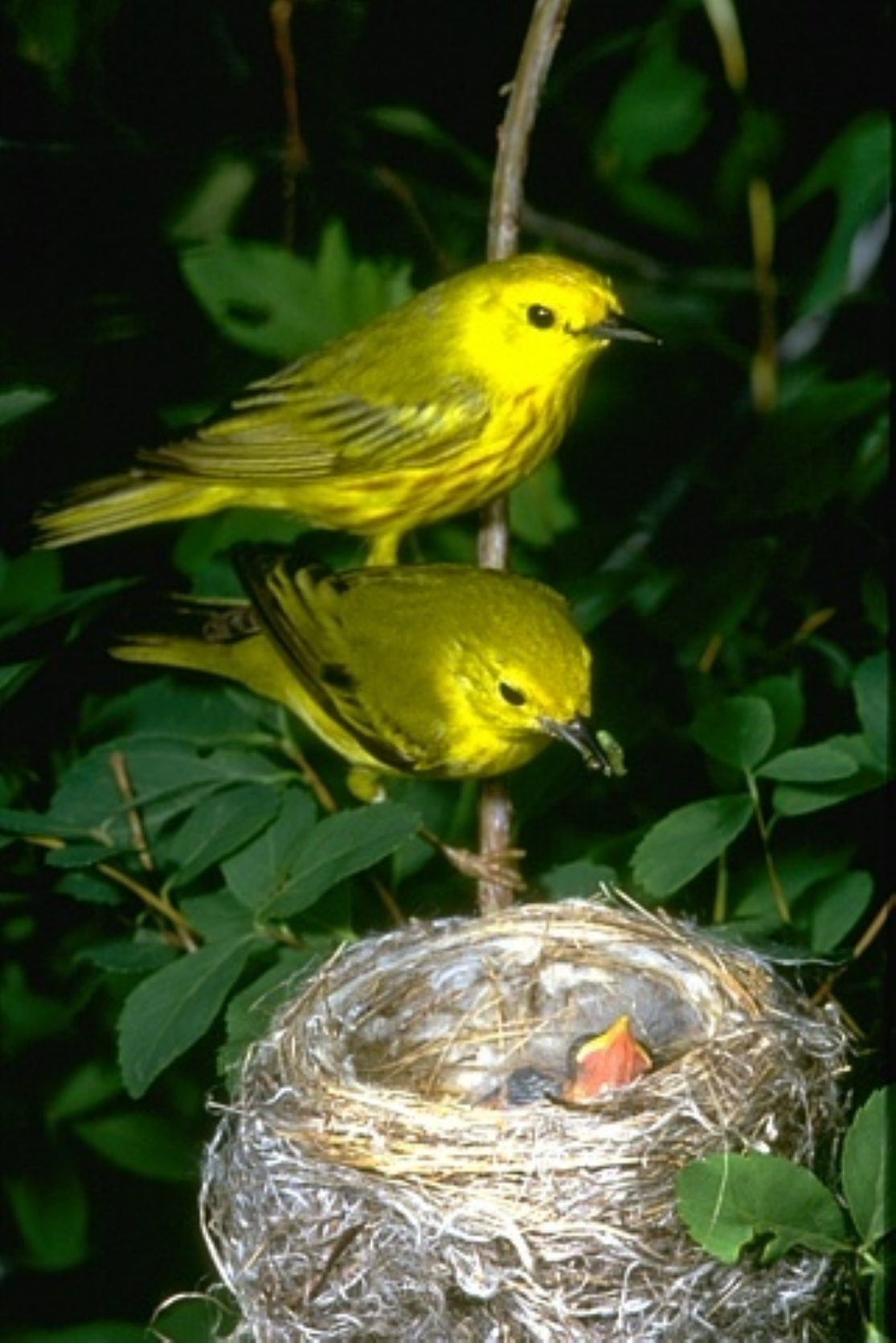}
  }
  \subfloat[Ground truth \#1]{
    \includegraphics[width=21mm]{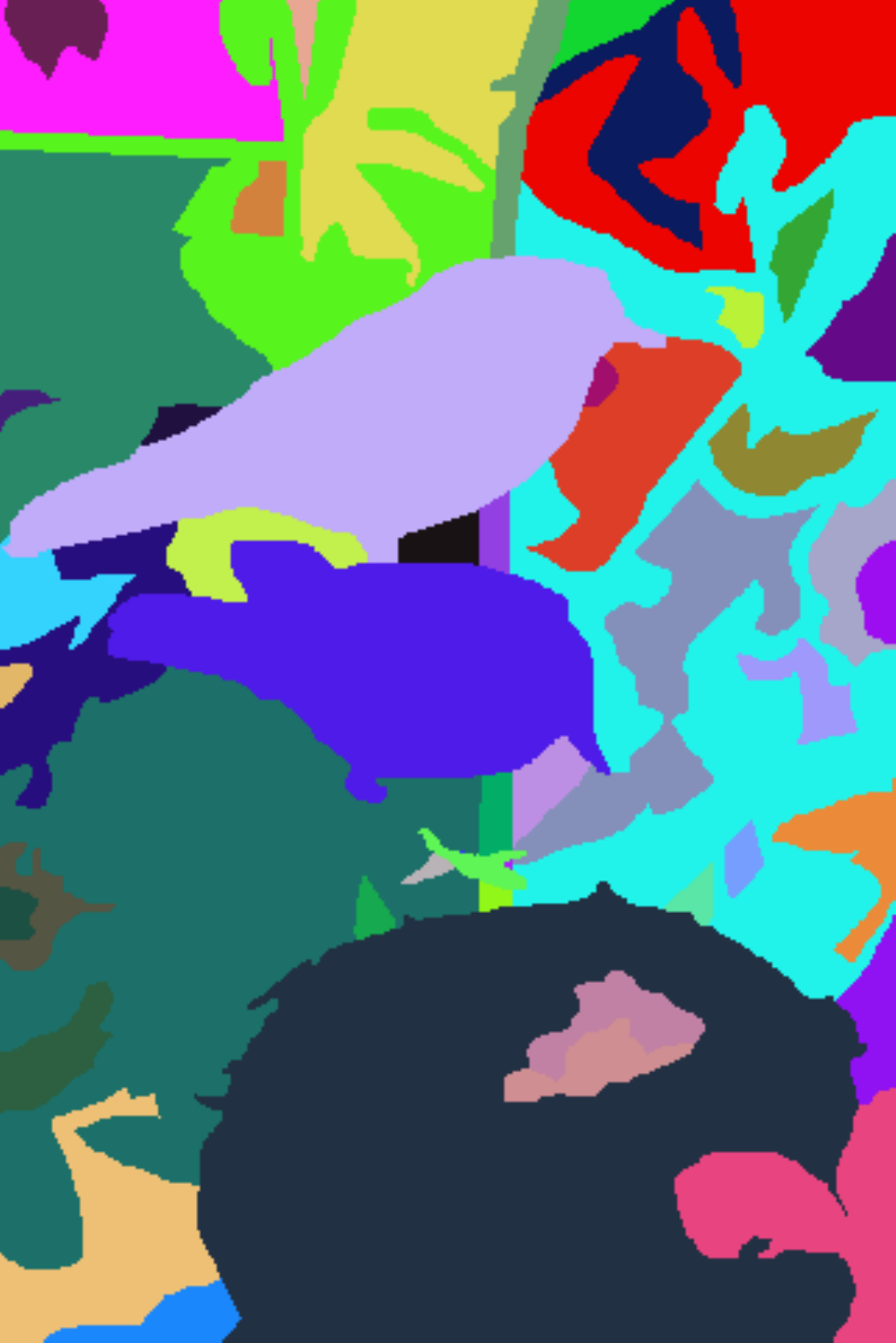}
    \label{fig:comparison_BSD500_gt01}
  }
  \subfloat[Ground truth \#2]{
    \includegraphics[width=21mm]{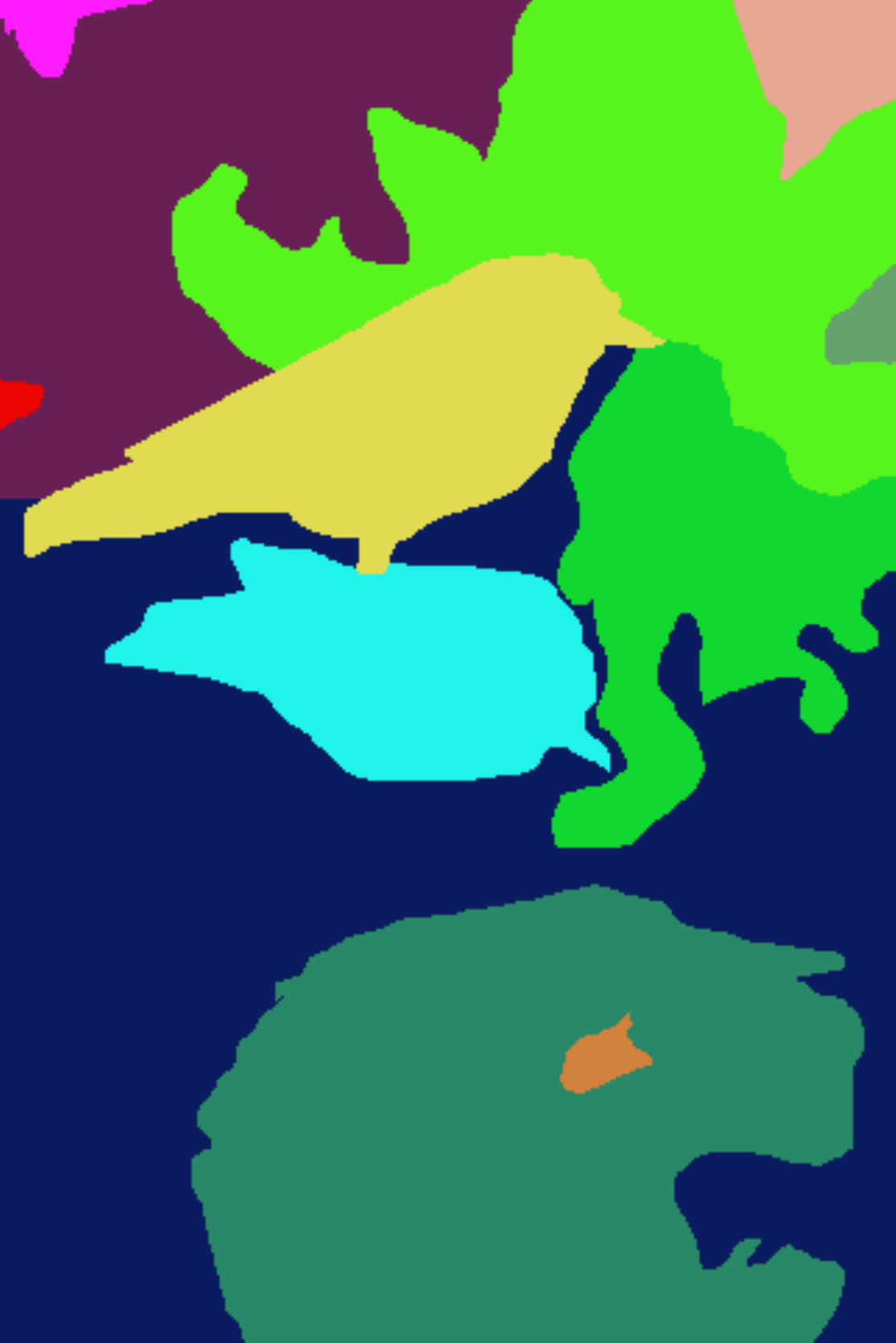}
    \label{fig:comparison_BSD500_gt02}
  }
  \subfloat[GS \cite{felzenszwalb2004efficient}]{
    \includegraphics[width=21mm]{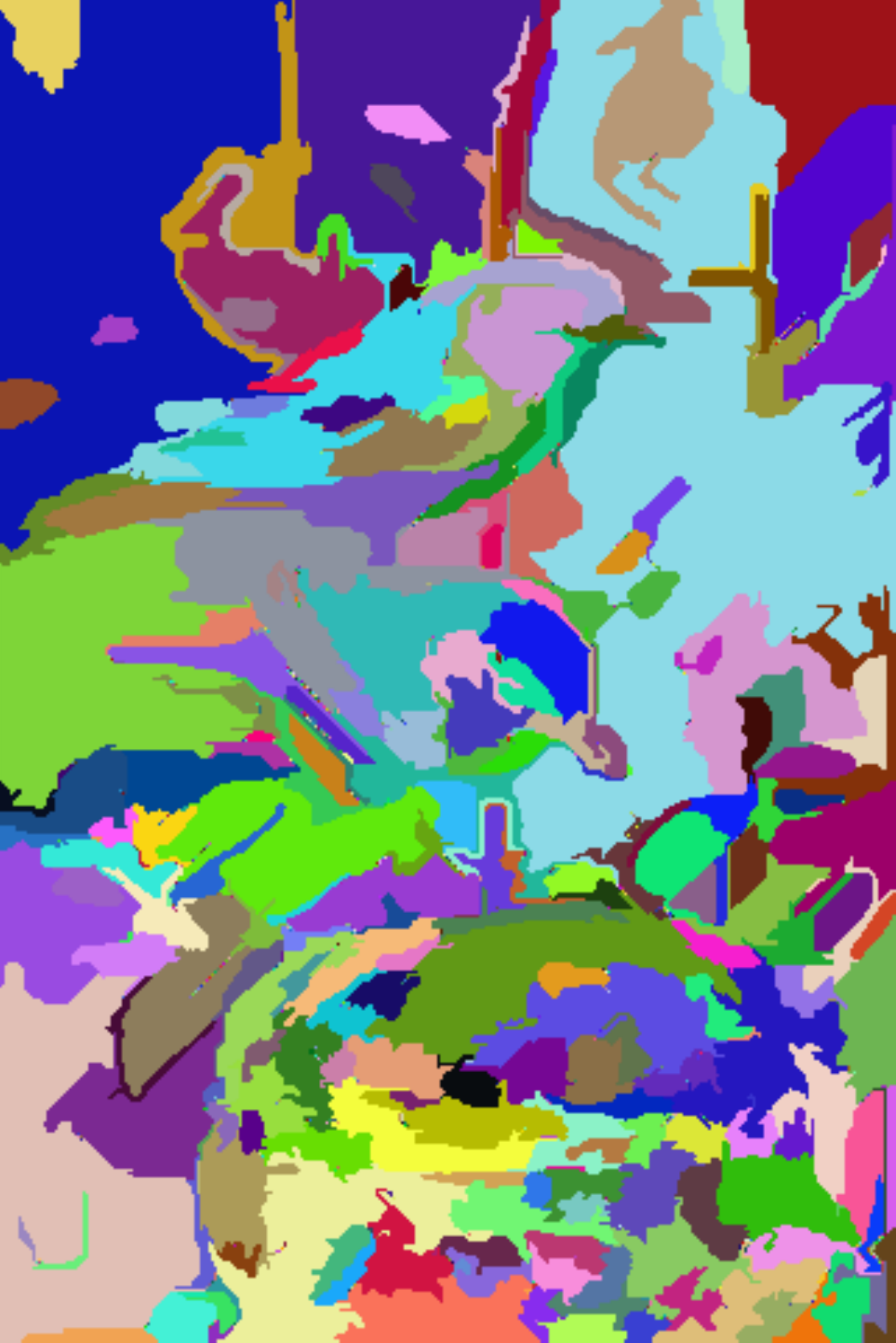}
  }
  \subfloat[$k$-means]{
    \includegraphics[width=21mm]{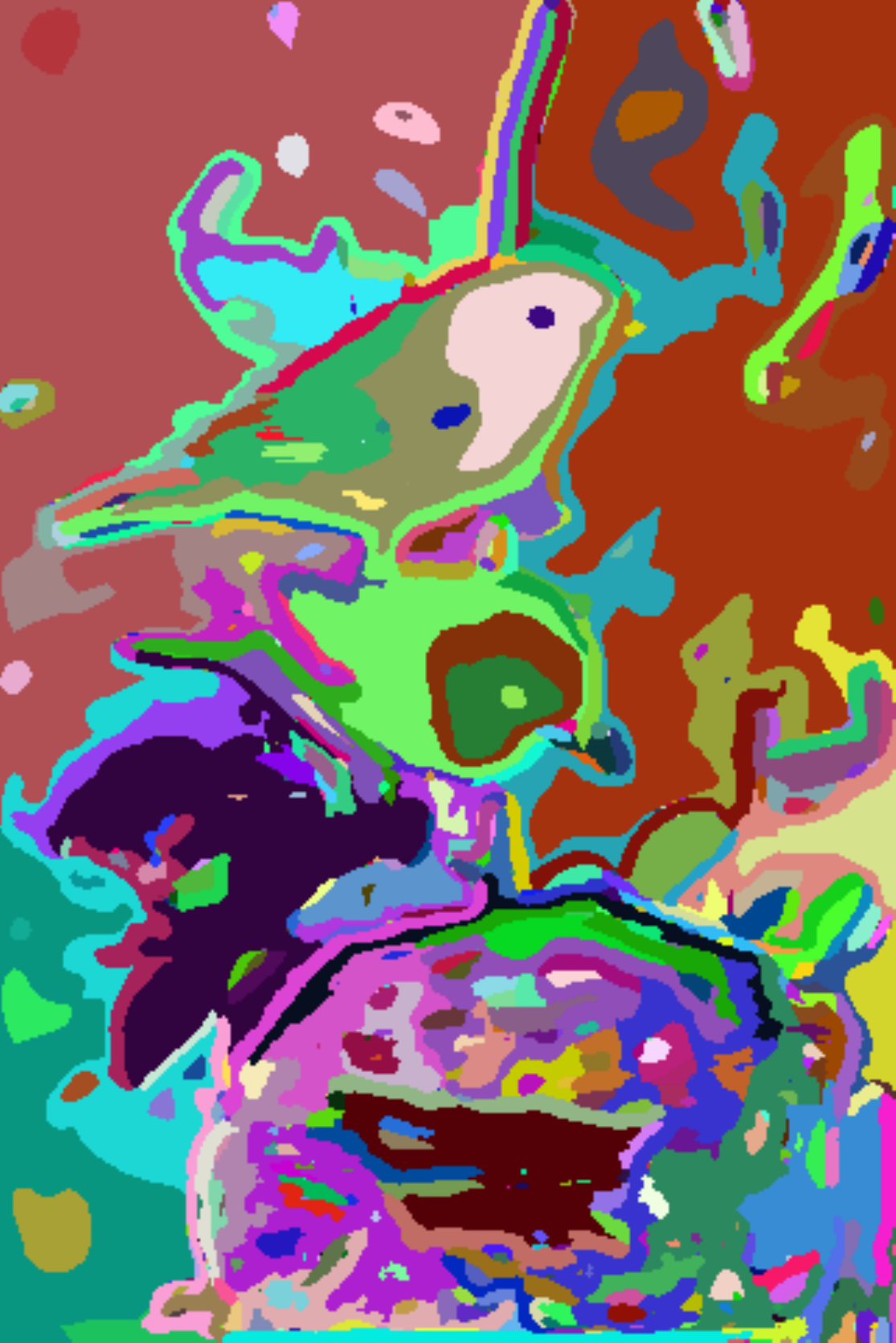}
  }
  \subfloat[IIC \cite{ji2019invariant}]{
    \includegraphics[width=21mm]{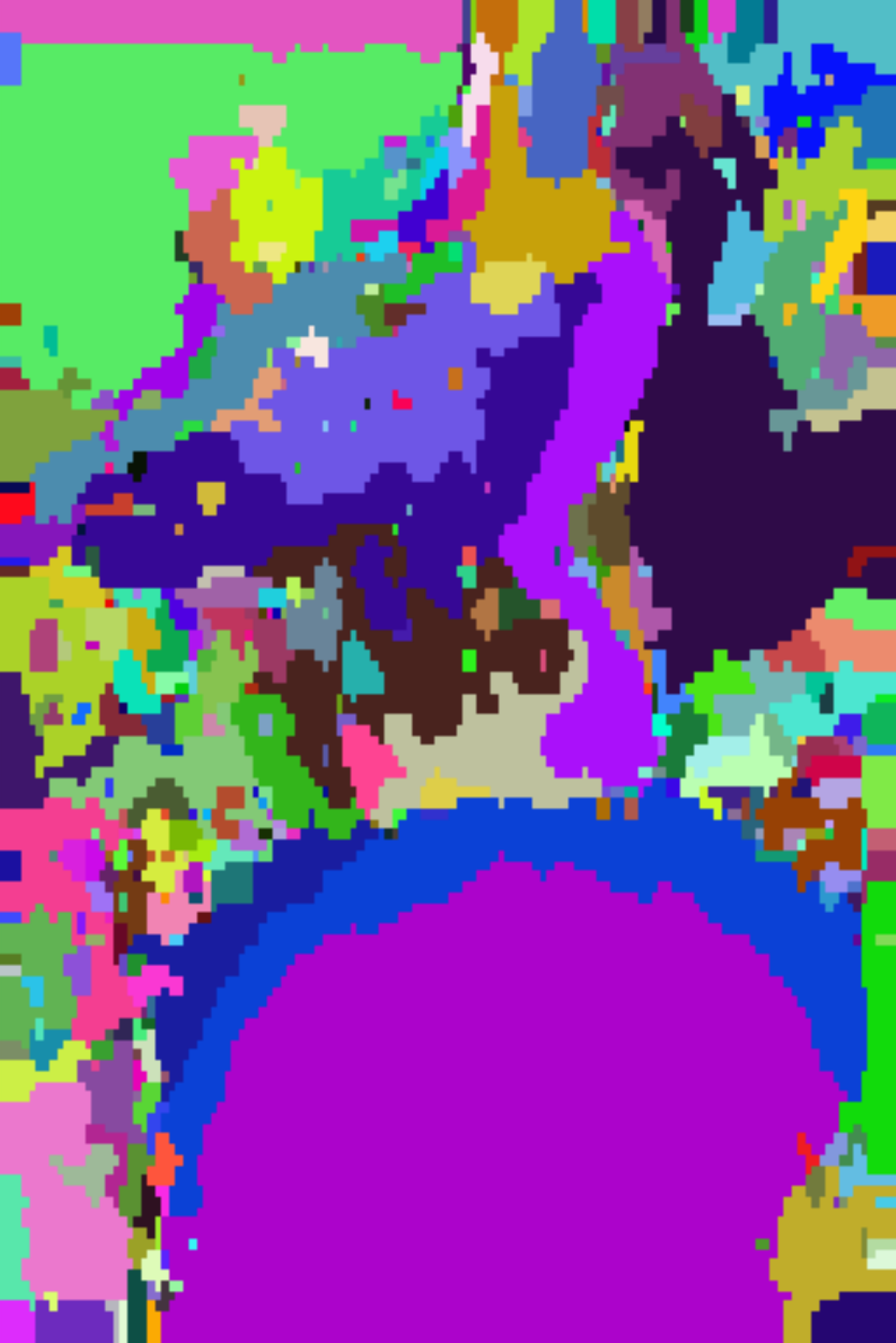}
  }
  \subfloat[w/ superpixels \cite{kanezaki2018unsupervised}]{
    \includegraphics[width=21mm]{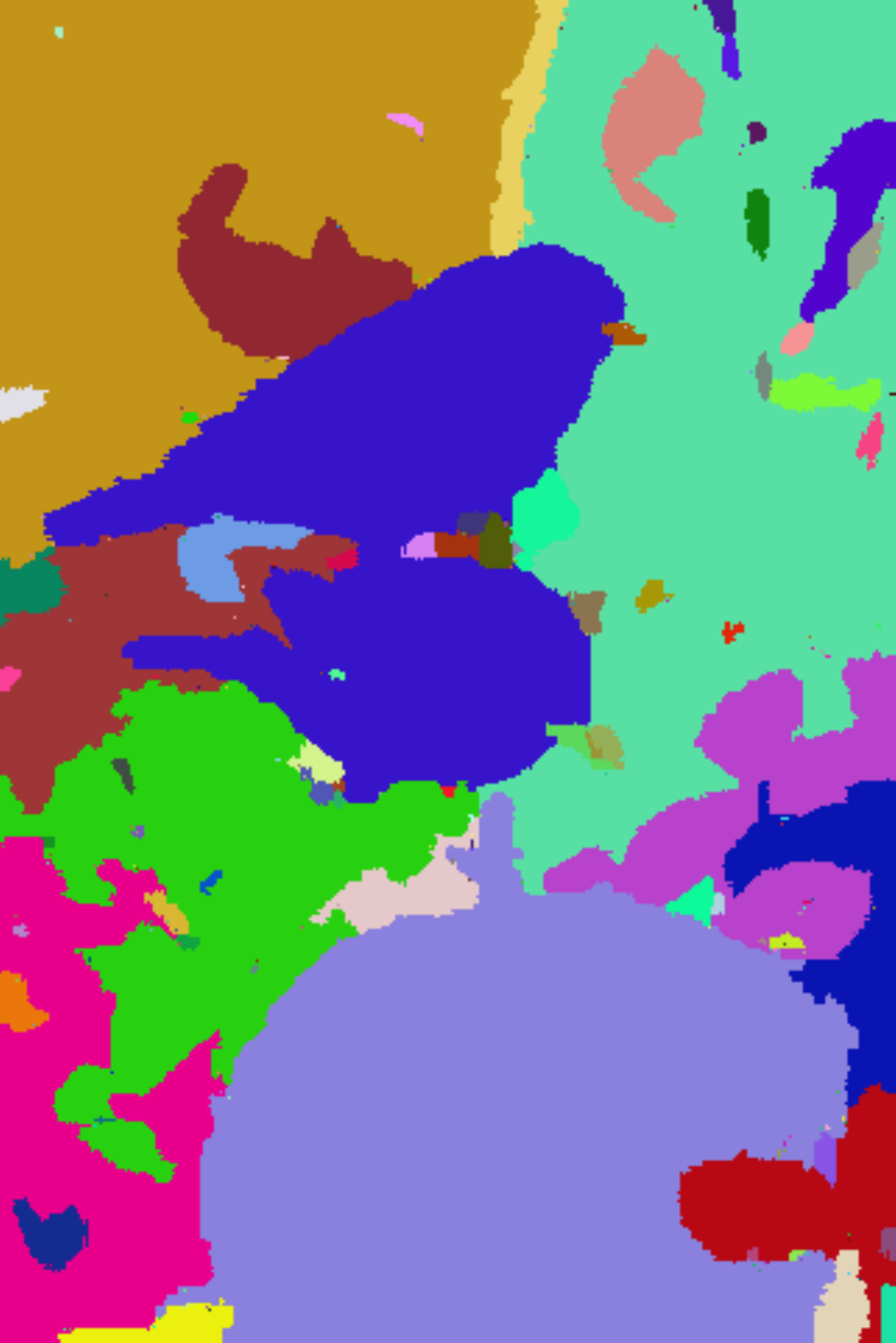}
  }
  \subfloat[w/ continuity loss]{
    \includegraphics[width=21mm]{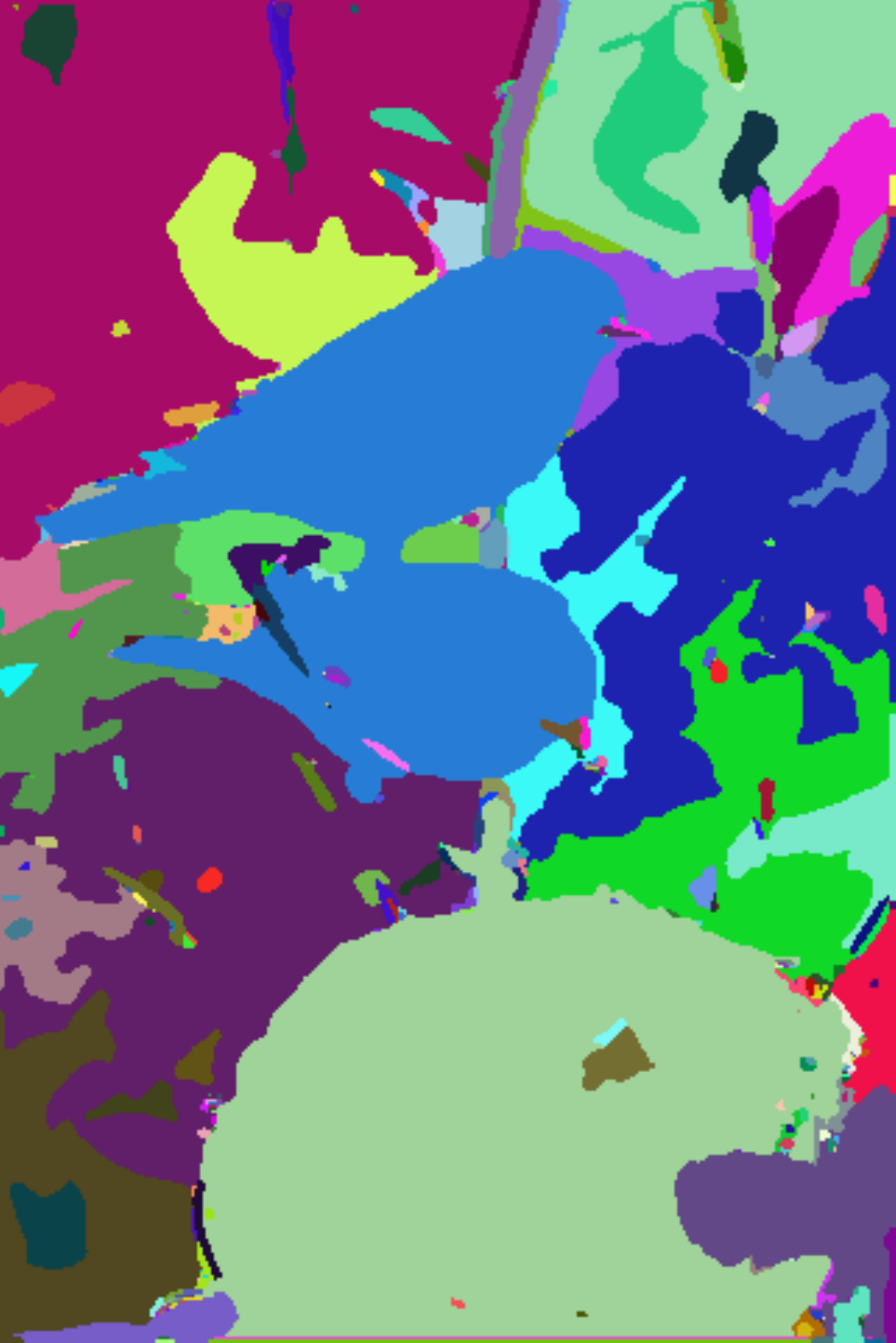}
  }
  \end{center}
  \caption{Comparison of unsupervised segmentation results on BSD500. Different segments are shown in different colors. In (b) and (c), two different ground truth segments of image (a) in BSD500 superpixel benchmark are shown. }
  \label{fig:comparison_BSD500}
\end{figure*} 

\begin{figure*}[t]
  \begin{center}
  \subfloat[IOU = 0.2]{
    \includegraphics[width=.33\textwidth]{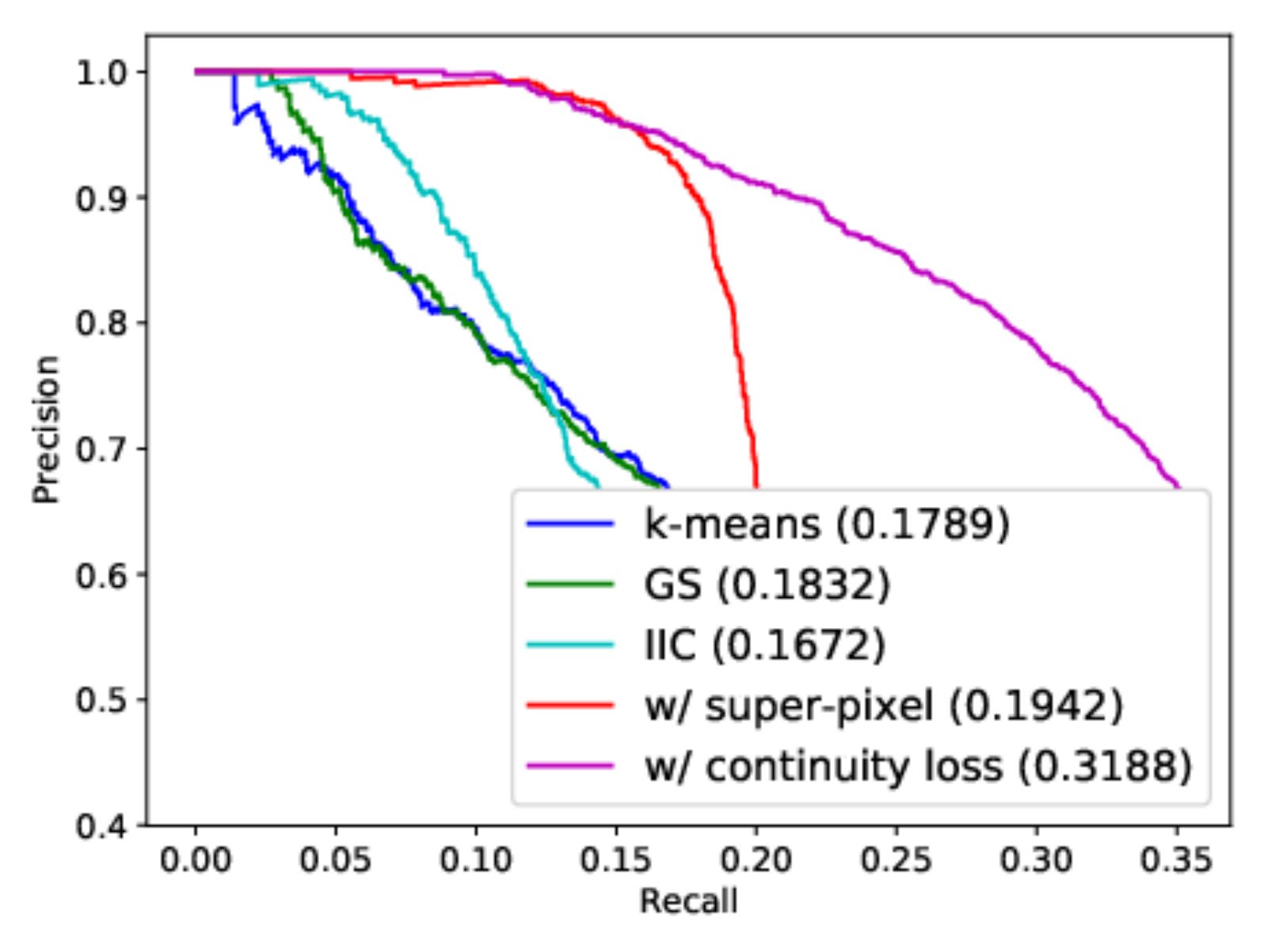}
  }
  \subfloat[IOU = 0.3]{
    \includegraphics[width=.33\textwidth]{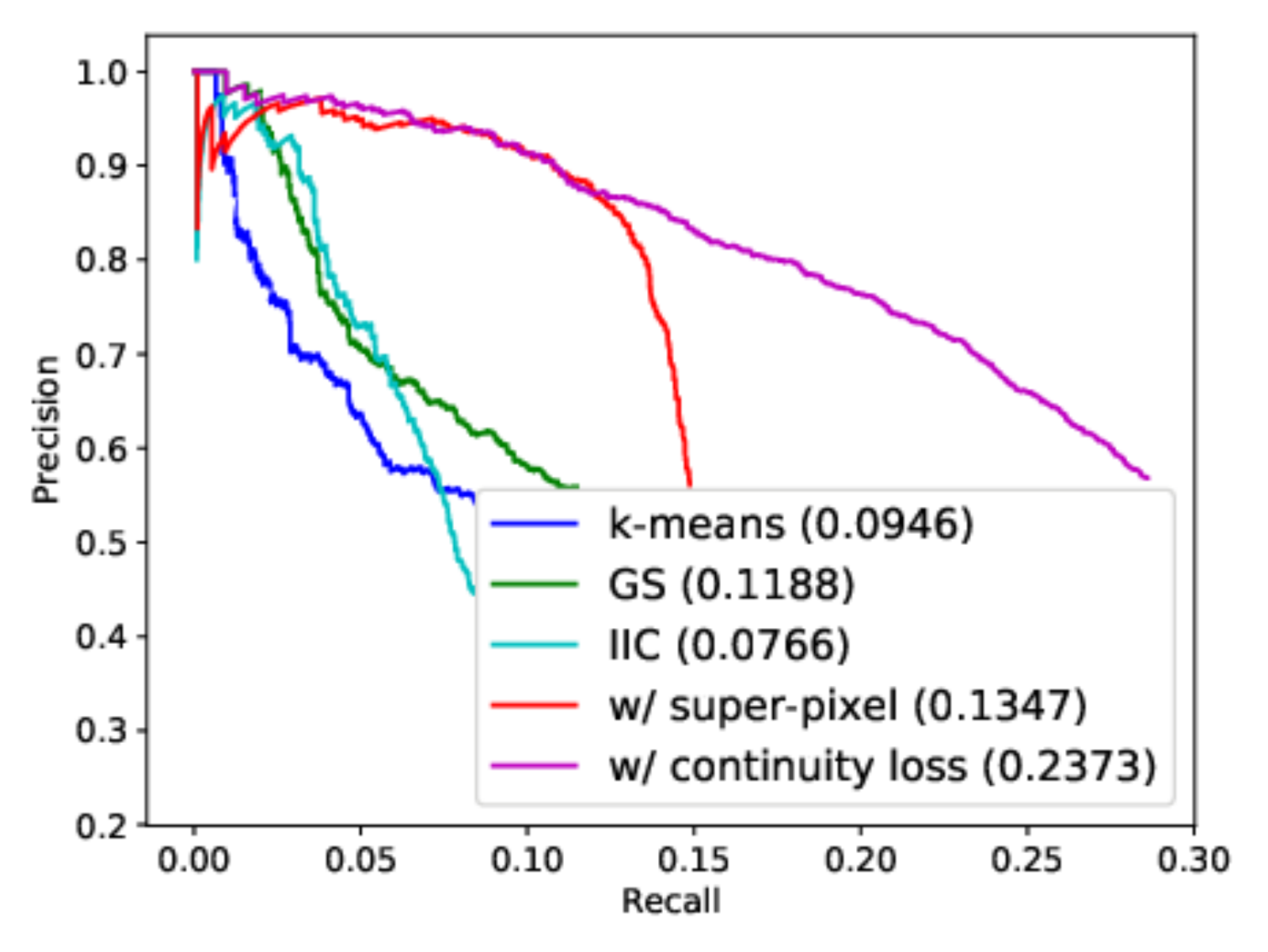}
  }
  \subfloat[IOU = 0.4]{
    \includegraphics[width=.33\textwidth]{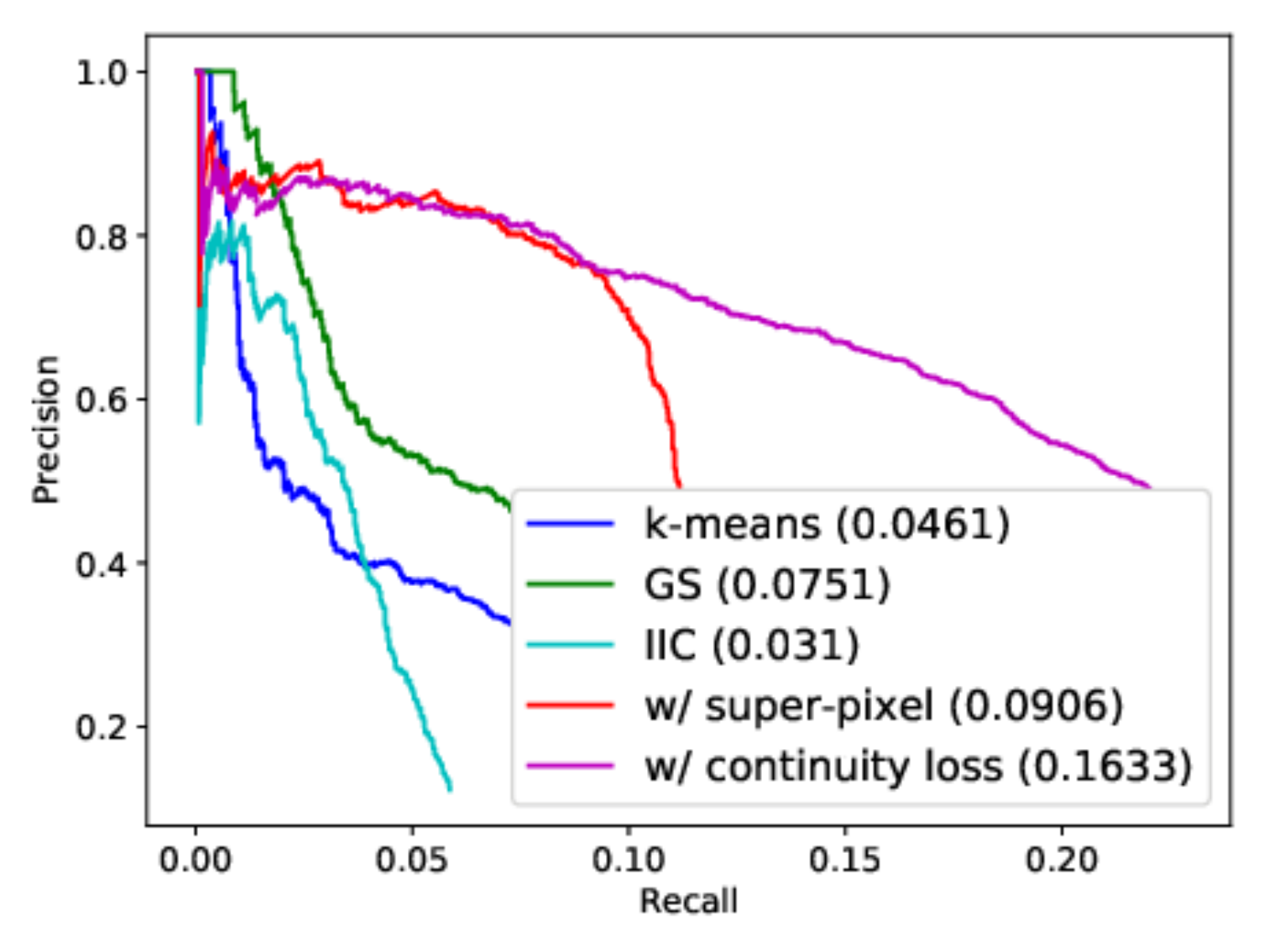}
  } \\
  \subfloat[IOU = 0.5]{
    \includegraphics[width=.33\textwidth]{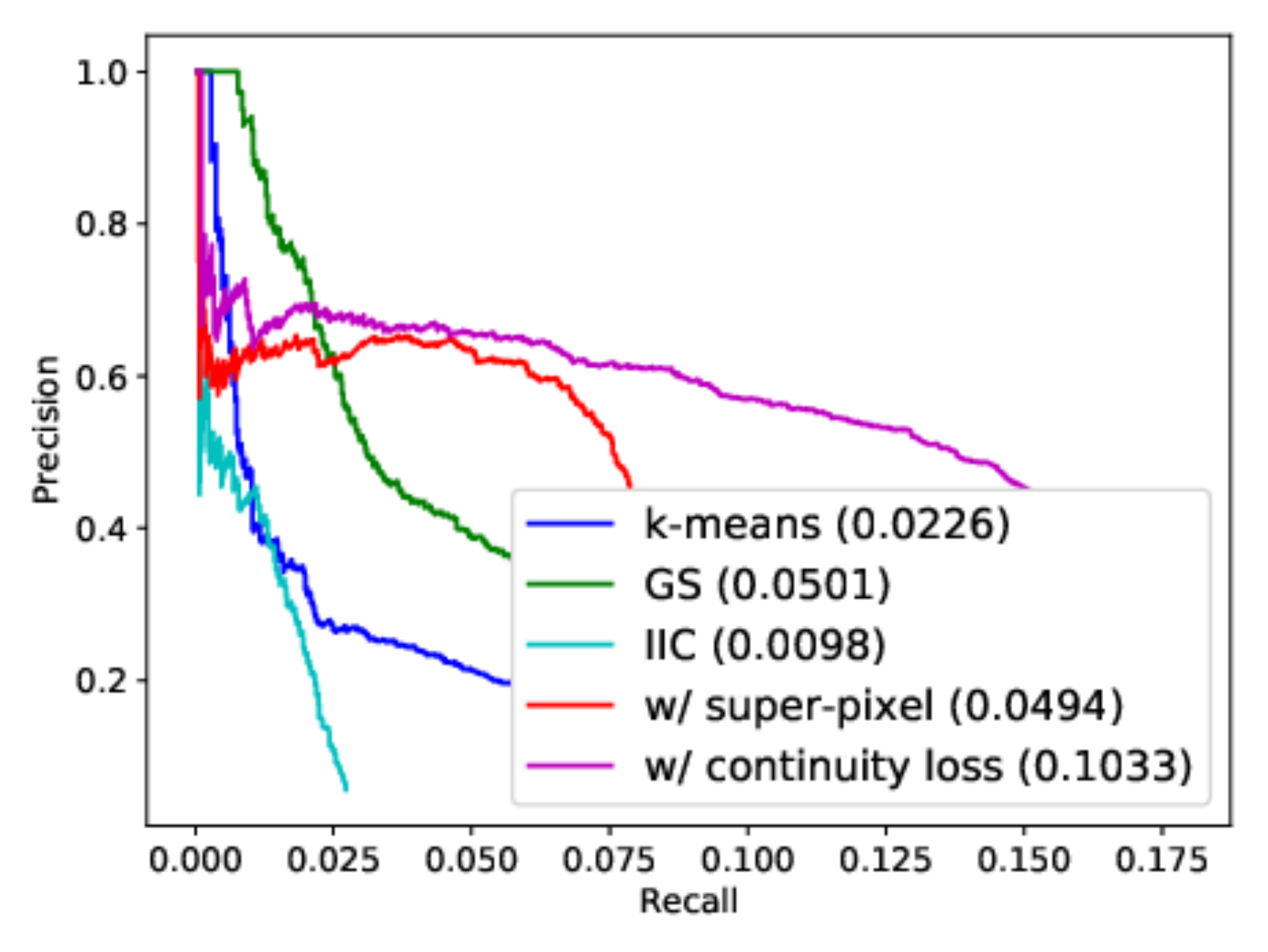}
  }
  \subfloat[IOU = 0.6]{
    \includegraphics[width=.33\textwidth]{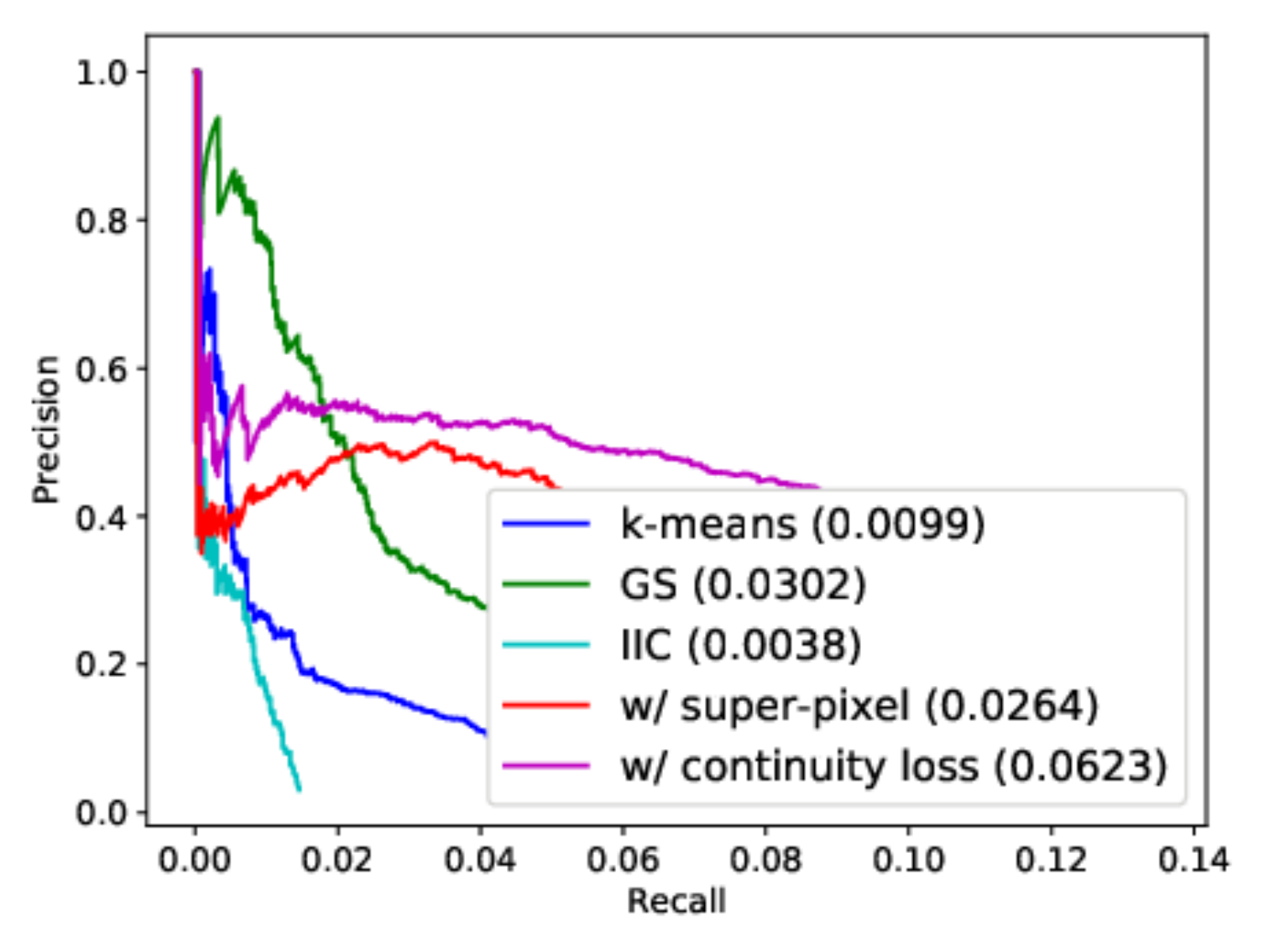}
  }
  \subfloat[IOU = 0.7]{
    \includegraphics[width=.33\textwidth]{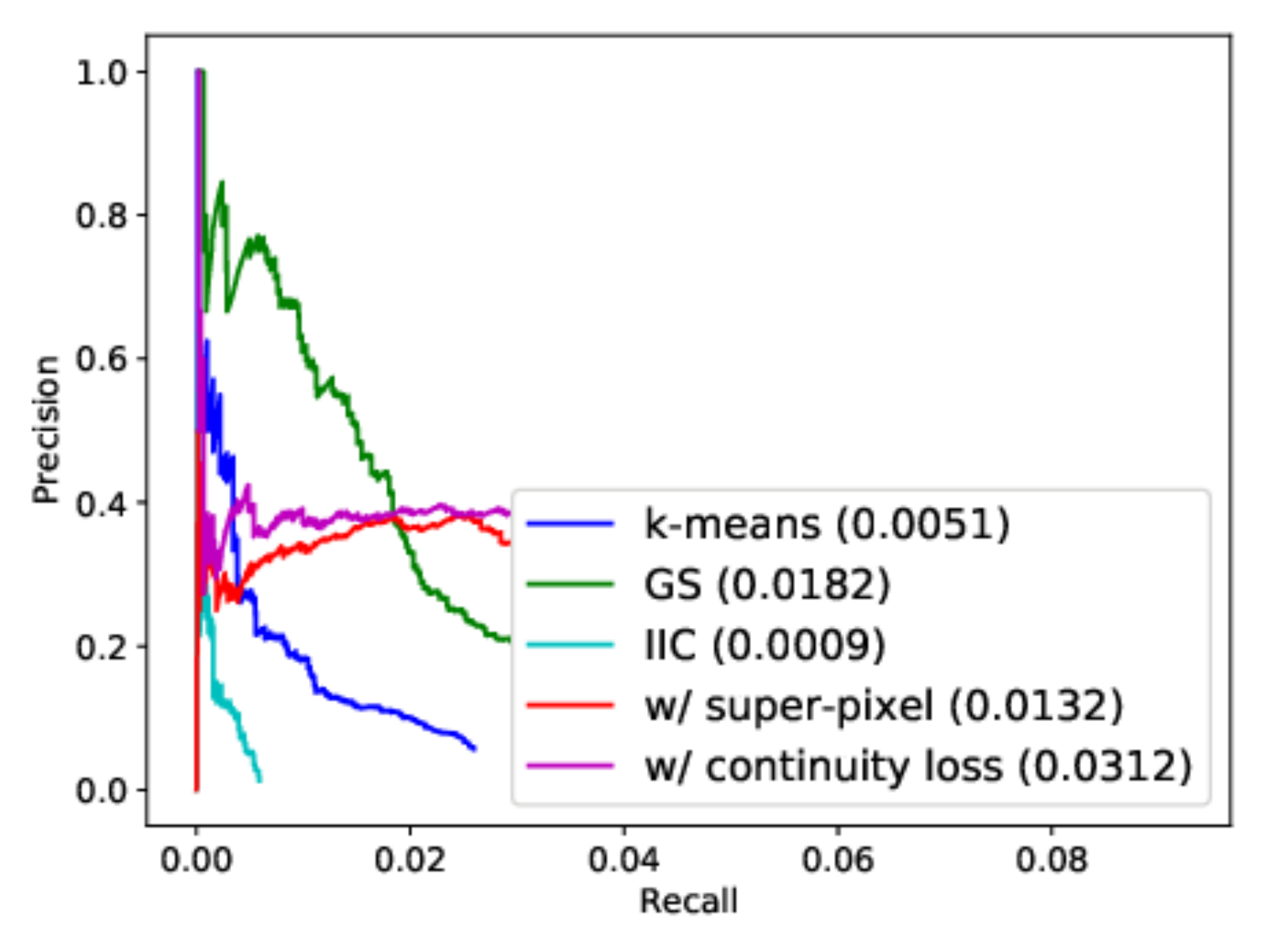}
  }
  \end{center}
  \caption{Precision-recall curves with different IOU thresholds for BSD500. The numbers in the legends represent the average precision scores of each method. }
  \label{fig:miou}
\end{figure*}

The effect of continuity loss on the validation dataset of PASCAL VOC 2012 segmentation benchmark \cite{everingham2015pascal} and Berkeley Segmentation Dataset and Benchmark (BSD500) \cite{amfm_pami2011} were evaluated.
\Fref{fig:comparison_parameter_con} shows examples of the segmentation results when $\mu$ was changed.
In case of \fref{fig:comparison_parameter_con_f}, the image was successfully segmented into sky, sea, rock, cattle, and beach regions.
However, the image was segmented in more detail with $\mu = 1$; for example, the beach was further segmented into sand and grass regions.
It is inferred that the optimal $\mu$ changes depending on the degree of detailing in the desired segmentation results.
\Tref{table:prs} shows the change in mIOU scores with respect to $\mu$ and $\nu$ variations on PASCAL VOC 2012 dataset \cite{everingham2015pascal}. The results show that $\mu=5$ is the best when applying to unsupervised segmentation and $\nu=0.5$ is the best for segmentation with user input.
It is also shown that the proposed method is more sensitive to $\nu$ than $\mu$.

\Tref{table:closs} shows comparative results of the unsupervised image segmentation on two benchmark datasets.
The $k$-means clustering and the graph-based segmentation method (GS) \cite{felzenszwalb2004efficient} were chosen as the comparative methods.
In case of GS, a gaussian filter with $\sigma=1$ was applied to smooth an input image slightly before computing the edge weights, in order to compensate for digitization artifacts. GS needs a threshold parameter to determine the granularity of segments. The threshold parameter effectively sets a scale of observation, in that a larger value causes a preference for larger components.
For the $k$-means clustering, the concatenation of RGB values in a $5 \times 5$ window were used for each pixel representation.
The connected components were extracted as segments from each cluster generated by $k$-means clustering and the proposed method.
The best $k$ for $k$-means clustering and threshold parameter $\tau$ for GS were experimentally determined from $\{2, 5, 8, 11, 14, 17, 20\}$ and $\{100, 500, 1000, 1500, 2000\}$, respectively.
For comparison with a cutting-edge method, we employed Invariant Information Clustering (IIC) \cite{ji2019invariant}.
We altered the number of output clusters and iterations as $\{2, 5, 8, 11, 14, 17, 20\}$ and $\{10, 20, 30, 40, 50\}$, respectively, and shows the best result among them.
As to other parameters, we used default values used for Microsoft COCO dataset in the official IIC code~\footnote{\url{https://github.com/xu-ji/IIC}}.

Examples of unsupervised image segmentation results on PASCAL VOC 2012 and BSD500 are shown in \fref{fig:result_image_con} and \fref{fig:comparison_BSD500}, respectively.
As shown in figure, the boundary lines of segments are smoother and more salient using the proposed method compared with those in our previous work \cite{kanezaki2018unsupervised}.
This improvement also leads to enhanced performance, which can be confirmed from \Tref{table:closs}.
There are several sets of ground truth segmentation for each image in the BSD500 superpixel benchmark.
Figures \ref{fig:comparison_BSD500_gt01} and \ref{fig:comparison_BSD500_gt02} show two different ground truth segments for an exemplar test image.
As shown in figure, the ground truth segments are labeled without certain object classes.
For evaluation, three groups for mIOU calculation were defined as: ``all'' using all ground truth files, ``fine'' using a single ground truth file per image that contains the largest number of segments, and ``coarse'' using the ground truth file that contains the smallest number of segments.
In this case, ``fine'' used \fref{fig:comparison_BSD500_gt01}, ``coarse'' used \fref{fig:comparison_BSD500_gt02}, and ``all'' used all the ground truth files including both of those for mIOU calculation.
According to \Tref{table:closs}, the proposed method achieved the best or the second best scores on PASCAL VOC 2012 and BSD500 datasets.
The proposed method was outperformed by GS on ``BSD500 all'' and ``BSD500 fine'' because the IOU values for small segments are dominant owing to the several small segments in the ground truth sets.
This in effect does not convey that the proposed method produced fewer accurate segments than GS.
To confirm this fact, the precision-recall curves in ``BSD500 all'' with an IOU threshold $0.2$, $0.3$, $0.4$, $0.5$, $0.6$, and $0.7$ in \fref{fig:miou} were also presented.
For this evaluation, we first sort all the estimated segments according to maximum IOU values between respective estimated segments and the ground truth segments.
The precision-recall curves in \fref{fig:miou} are drawn by counting an estimated segment as a true positive when the maximum IOU between the estimated segment and the ground truth segments exceeds a threshold.
The number of true positive segments reduces when the threshold increases, which causes different average precision scores in the plots in \fref{fig:miou}.

The proposed method w/ continuity loss achieved the best average precision scores and our previous method w/ superpixels~\cite{kanezaki2018unsupervised} achieved the second best average precision scores for all the cases in \fref{fig:miou}.

To confirm the effectiveness of each element of the proposed method, the ablation study was performed with PASCAL VOC 2012 and BSD500. \Tref{table:abl study} shows the results regarding the presence and absence of $L_{\rm{con}}$ and the batch normalization of the response map.
The experimental results show that the batch normalization process consistently and considerably improves the performance on all the datasets.
Even though the effect of $L_{\rm{con}}$ alone is marginal, it gives a solid improvement when used together with the batch normalization.
This indicates the importance of the three criteria introduced in \sref{sec:introduction}.

\subsection{Segmentation with scribbles as user input}
\label{sec:weaklyexp}

%
\begin{figure*}[t]
  \begin{center}
    \includegraphics[width=\linewidth]{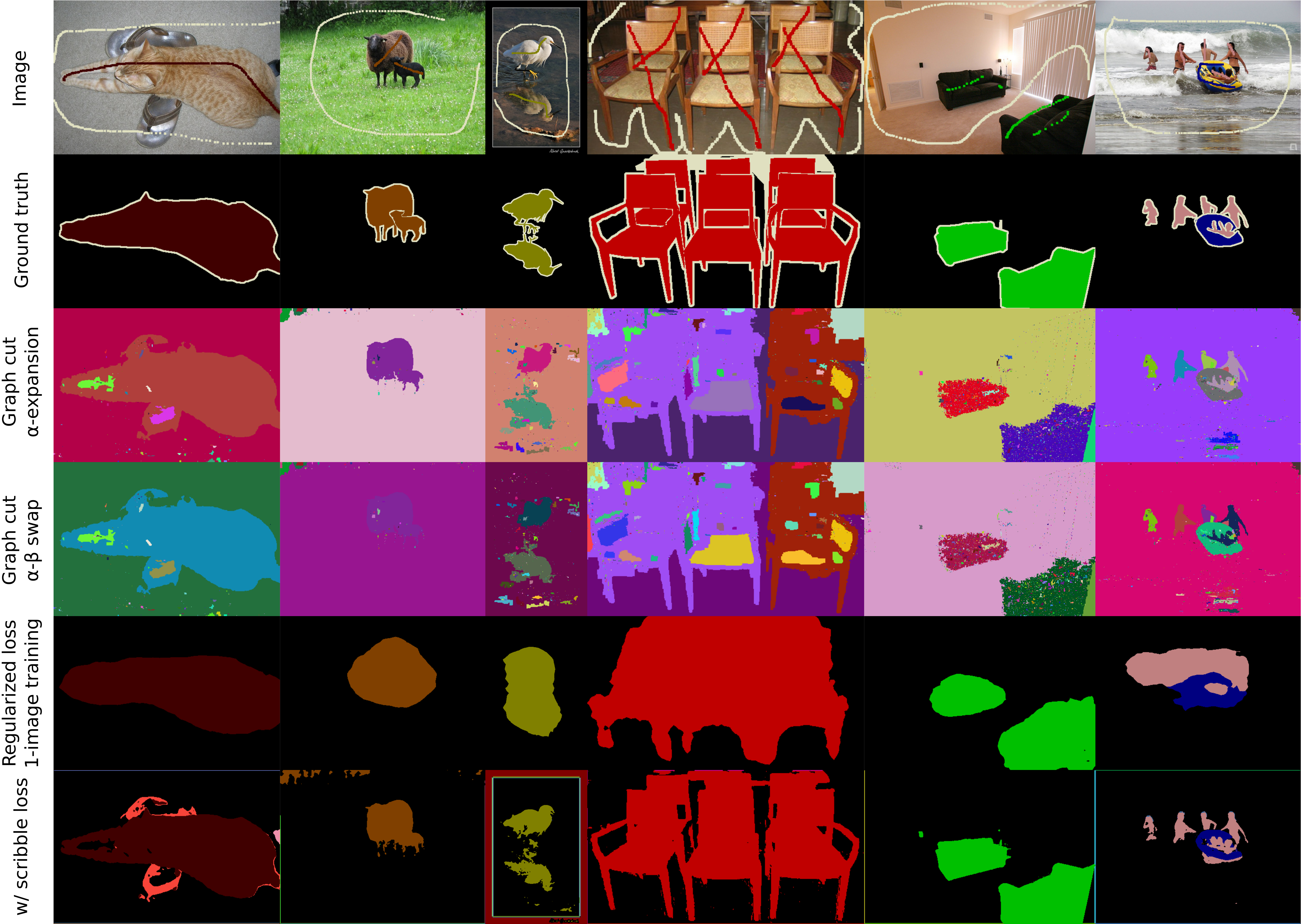}
  \end{center}
  \caption{Comparison of segmentation results with user input. We used DeepLab-ResNet-101 for the base architecture of ``Regularized loss 1-image training''. ``Image'' row shows input images including scribbles (user input), which are made bold for the purpose of visualization. Different segments are shown in different colors.
  }
  \label{fig:result_image}
\end{figure*} 

The effect of the proposed method was tested for image segmentation with the user input on the validation dataset of PASCAL VOC 2012 segmentation benchmark \cite{everingham2015pascal}.
We let $\nu = 0.5$ in \eqref{eq:entireloss} in this experiment.
The scribble information was used for the test images given in \cite{lin2016scribblesup} as the user input.
For comparison, graph cut \cite{boykov1999fast}, graph cut $\alpha$-expansion \cite{boykov1999fast}, graph cut $\alpha$-$\beta$ swap \cite{boykov1999fast}, and regularized loss \cite{tang2018regularized} were employed.
In graph cut, Gaussian Mixture Model (GMM) was used for modeling foreground and background of an image. A graph is constructed from pixel distributions modeled by GMM. At this time, scribbled pixels are fixed to their scribbled labels which are foreground or background. In the generated graph, a node is defined as a pixel, whereas the weight of an edge connecting nodes is defined by a probability to be foreground or background. Thereafter, the graph is divided by energy minimization into the two groups: foreground and background.
The vanilla graph cut is an algorithm for segmenting the foreground and the background and it does not support the multi-labels case.
Therefore, in this study, a graph cut was performed multiple times where each scribble is regarded as the foreground each time, and subsequently all the extracted segments were used for calculating the mIOU.
To compare the performance, $\alpha$-expansion and $\alpha$-$\beta$ swap (introduced in \sref{sec:related_work_input}), as well as regularized loss \cite{tang2018regularized} were tested.
Regularized loss \cite{tang2018regularized} is a weakly-supervised method of segmentation using a training data set and additional scribble information.
%
In order to unify the experiment conditions, one image from the validation dataset of PASCAL VOC 2012 is used for the network training with the scribble information.
The output in the final iteration for the image after completion of training was regarded as the segmentation result of the image.
After that, the network weight is initialized, and then the process is repeated for the next image.
This process was repeated individually for all the test images in the validation dataset of PASCAL VOC 2012.
This process was defined as ``Regularized loss 1-image training''.
We tested two base architectures for ``Regularized loss 1-image training'': DeepLab-largeFOV and DeepLab-ResNet-101.

\begin{table*}
 \caption{Comparison of the number of parameters, computation time, and mIOU for segmentation with user input. }
 \begin{center}
 {\normalsize
  \begin{tabular}{lrrr}
  \toprule
     Method & \# parameters & Time (sec.) & mIOU \\ \midrule
    Graph cut \cite{boykov1999fast} & - & 1.47 &  0.2965 \\
    Graph cut $\alpha$-expansion \cite{boykov1999fast} & - & 0.81 & 0.5509 \\
    Graph cut $\alpha$-$\beta$ swap \cite{boykov1999fast} & - & 0.77 &  0.5524 \\
    Regularized loss \cite{tang2018regularized} 1-image training (DeepLab-largeFOV) & 20,499,136 & 42 & 0.5790 \\
    Regularized loss \cite{tang2018regularized} 1-image training (DeepLab-ResNet-101) & 132,145,344 & 414 & 0.6064 \\
    Proposed method w/ scribble loss & 103,600 & 20 & {\bf 0.6174} \\ \bottomrule 
  \end{tabular}
 }
 \end{center}
 \label{table:sloss}
\end{table*}

\begin{figure*}[t]
  \begin{center}
  \subfloat[Reference images]{
        \includegraphics[width=150mm]{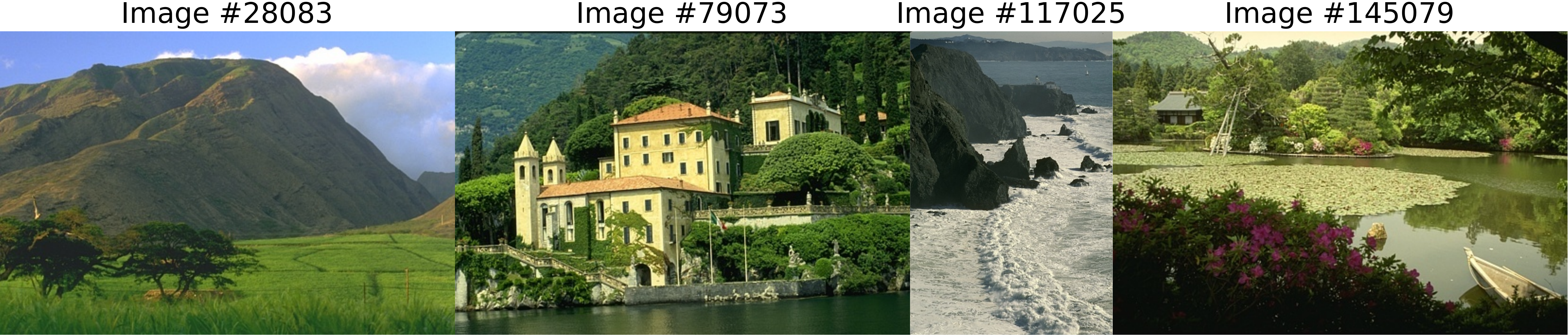}
        \label{fig:BSDtrain}
  } \\
  \subfloat[Test results]{
        \includegraphics[width=150mm]{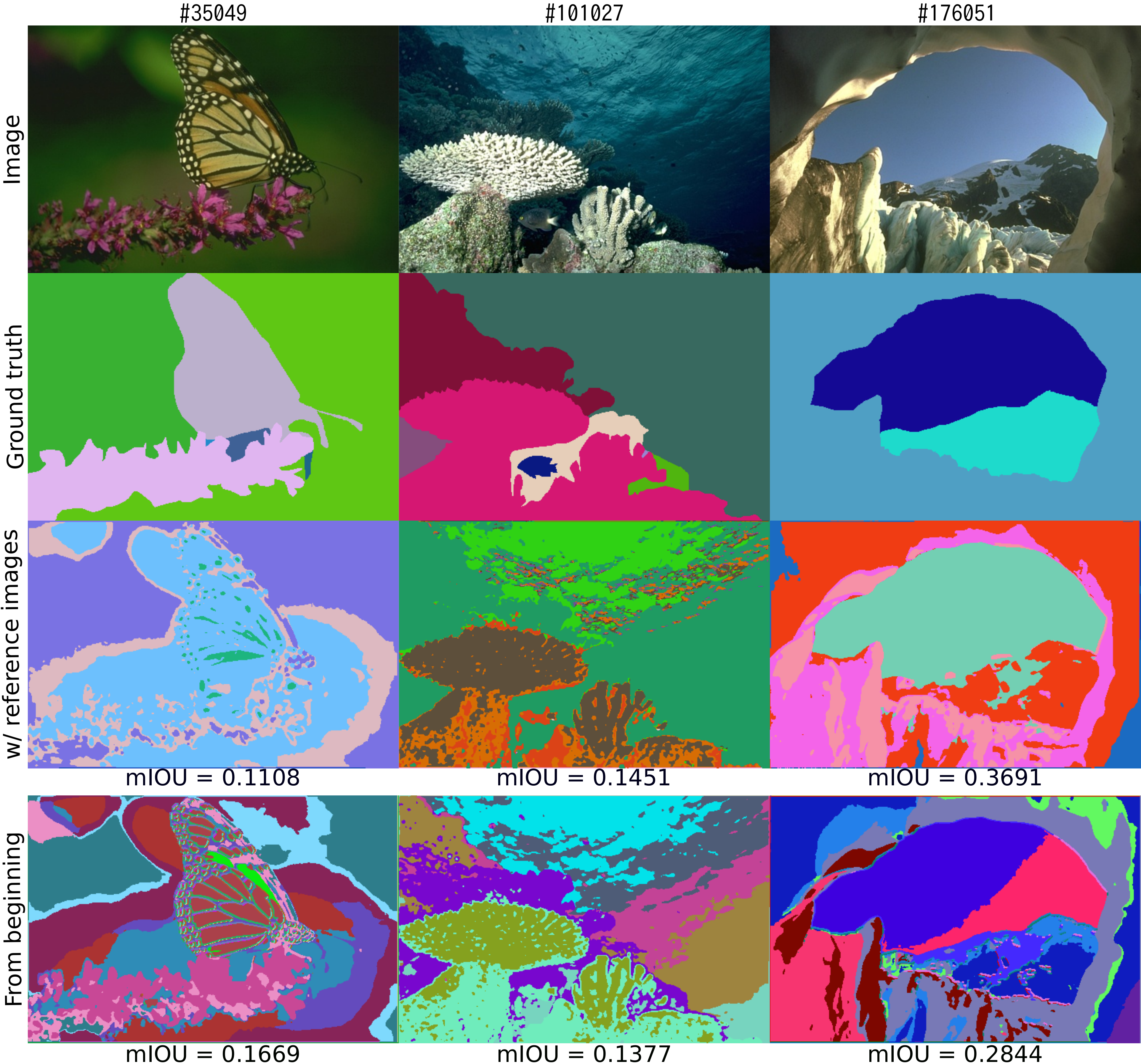}
        \label{fig:BSDtest}
  } \\

  \end{center}
  \caption{Results of segmentation with reference images on BSD500.
    Different segments are shown in different colors.}
  \label{fig:BSD}
\end{figure*}

Exemplar segmentation results are shown in \fref{fig:result_image}.
It was observed that the proposed method is more stable than the graph-based methods.
Relatively rougher segments of objects are detected by ``Regularized loss 1-image training'', whereas the boundaries of segmented areas of the proposed method are more accurate.
The quantitative evaluation in \Tref{table:sloss} shows that the proposed method achieved the best mIOU score. 
In addition to outperforming than ``Regularized loss 1-image training'' with the DeepLab-ResNet-101 architecture, the proposed method is effective in three folds.
First, the proposed method uses a small network where the number of parameters is $1,000$ times less than DeepLab-ResNet-101.
Owing to the smallness of the architecture, the proposed method converges $20$ times faster than ``Regularized loss 1-image training'' with the DeepLab-ResNet-101 architecture.
Finally, the proposed method initializes the network with random weights and thus requires no pre-trained weights.
In contrast, ``Regularized loss 1-image training'' requires the weights pre-trained on \eg, the ImageNet dataset\footnote{We used the pre-trained weights downloaded from \url{http://liangchiehchen.com/projects/Init\%20Models.html} and \url{https://github.com/KaimingHe/deep-residual-networks} for DeepLab-largeFOV and DeepLab-ResNet-101, respectively.} for initialization.
Notably, we found that ``Regularized loss 1-image training'' both with the DeepLab-ResNet-101 and DeepLab-largeFOV architectures failed to train the weights from random states in our experiment.

\subsection{Unsupervised segmentation with reference images}
\label{sec:fixweight}

\begin{figure*}[t]
  \begin{center}
    \includegraphics[width=\linewidth]{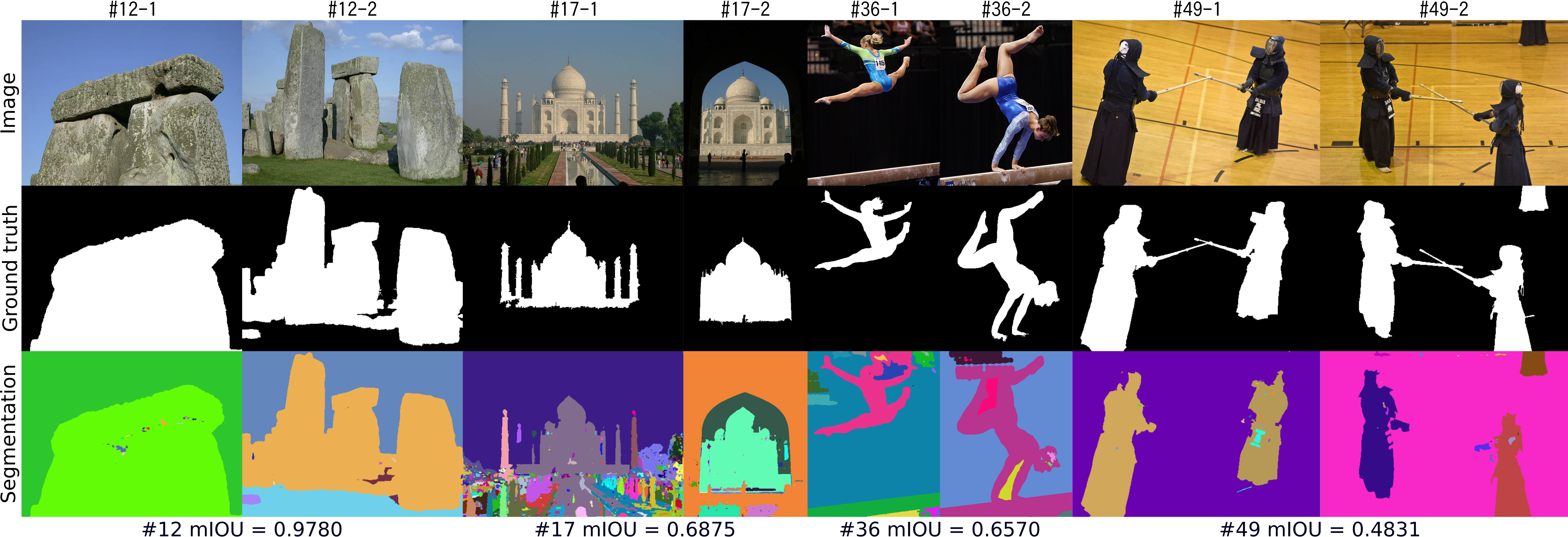}
  \end{center}
  \caption{Results of segmentation with reference images on iCoseg. ``$\#n$-$m$'' denotes the $m$th test image in the group whose ID is $n$.
    Different segments are shown in different colors.}
  \label{fig:icoseg}
\end{figure*} 

\begin{figure*}[t]
  \begin{center}
    \includegraphics[width=\linewidth]{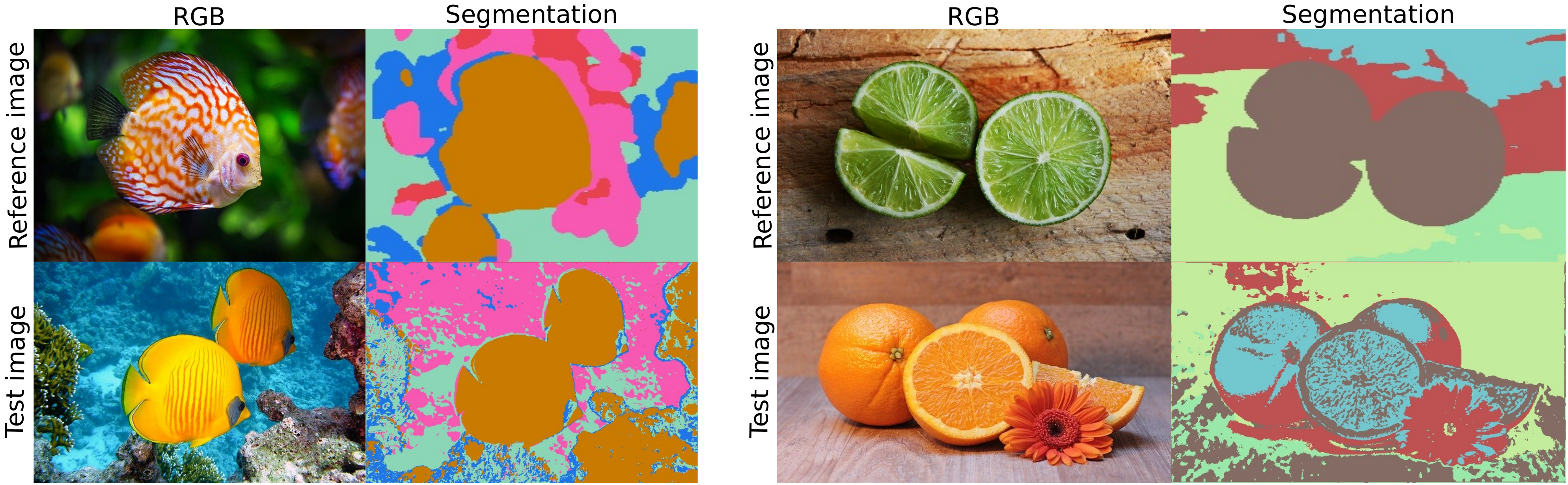}
  \end{center}
  \caption{Results of segmentation with a single reference image.
      Different segments are shown in different colors.
      The results imply that similar objects with somewhat similar colors are assigned the same label (see fishes in the left case), although otherwise not (see oranges in the right case).
      Images are from pixabay~\cite{pixabay}.
  }
  \label{fig:addit_ref}
\end{figure*} 

\begin{figure*}[t]
  \begin{center}
    \includegraphics[width=\linewidth]{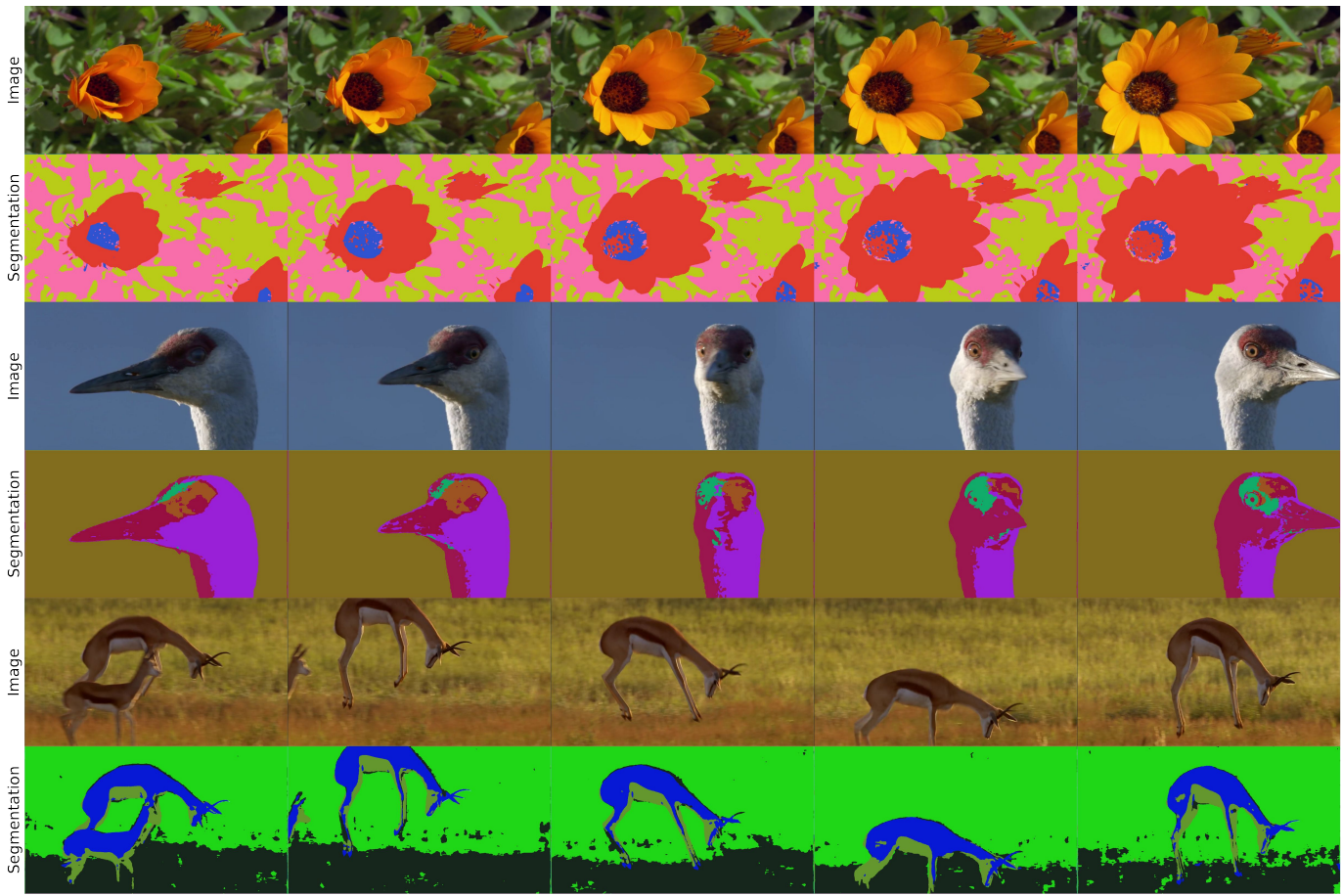}
  \end{center}
  \caption{Results of segmentation for sequential images. For each case, a network is trained with only the leftmost image in the respective row. Different segments are shown in different colors and the time flow is from left to right. The images are from ``BBC Earth, Nature Makes You Happy.''~\cite{BBC}
  }
  \label{fig:videoseg}
\end{figure*}

Supervised learning generally learns from training data and evaluates the performance using test data.
Therefore, the network can obtain segmentation results by processing the test images with the (fixed) learned weights.
In contrast, as the proposed method is completely unsupervised learning, it is necessary to learn the network weights every time the test image is input in order to obtain the segmentation results.
Further, an unsupervised segmentation experiment was conducted with reference images. 
The effectiveness of the networks of fixed weights trained on several images as reference was evaluated for unseen test images.
The BSD500 and the iCoseg \cite{batra2010icoseg} datasets were employed for the experiment. 

The proposed method was trained with four images in BSD500 shown in \fref{fig:BSDtrain}.
In the training phase, the network was updated once for each reference image.
After training, the network weights were fixed and the other three images shown in the top row in \fref{fig:BSDtest} were segmented.
The reference images and the test images were arbitrarily selected from different scenes in the nature category.
The segmentation results are shown in the two bottom rows in \fref{fig:BSDtest}.
The phrase ``from beginning'' in \fref{fig:BSDtest} means that an image is segmented with the proposed method where the weights of a network are trained for each test image from scratch.
As shown in \fref{fig:BSDtest}, the segmentation results ``w/ reference images'' were more detailed than ``from beginning''.
This is because ``from beginning'' integrates clusters under the influence of the continuity loss when training the target image.
According to \fref{fig:BSDtest}, ``w/ reference images'' showed acceptable segmentation performance compared with ``from beginning''.
The method ``w/ reference images'' only takes under $0.02s$ for the segmentation of each image, whereas the ``from beginning'' method takes approximately $20s$ under GPU calculation in GeForce GTX TITAN X to get the segmentation results.
The proposed method with four groups in iCoseg (ID: 12, 17, 36, 49) were also trained.
As iCoseg does not distinguish between the training and test data, two images from the group were randomly selected for testing.
Further, the proposed method was trained on the images in the group excluding the sampled test images.
The segmentation results are illustrated in \fref{fig:icoseg}.
Therefore, it was concluded that it is possible to segment unknown images with unsupervised trained weights on reference images, 
provided that the images are somewhat similar to the reference images (\eg, when they belong to the same category).

We also conducted an experiment to segment an image using a single reference image. \Fref{fig:addit_ref} shows the segmentation results of the test and reference images.
Even though the segmentation result for a test image is not as appropriate as that of a reference image, a sufficient segmentation result is obtained.
We can see different levels of quality in these two cases: fishes in the left case are successfully assigned the same label, whereas oranges in the right case are differentiated.
It implies that similar objects with somewhat similar colors are assigned the same label.

In the experiments thus far, it was found that the proposed method can be trained from several reference images and effective to similar-different images. 
Therefore, another application for videos was introduced.
Video data generally contains information connected in a time series.
Hence, video segmentation can be accomplished by training only a part of all the frames using the proposed method.
\Fref{fig:videoseg} shows examples of segmentation results when video data was input in the proposed method.
The proposed method trained a network only with the leftmost image in a respective row in \fref{fig:videoseg}.
It was observed that most of the segments obtained from other images were successfully matched to the same segments in the leftmost image.
Consequently, it was demonstrated that even the video data without ground truth can be segmented with the proposed method efficiently using only a single frame as a reference. 
This result indicates that the proposed method, which aims unsupervised learning of image segmentation, can be extended to unsupervised learning of video segmentation. By using the first frame of the video as a reference and segmenting other frames, the segmentation task can be accelerated. In addition, the segmentation of the full target video can also be improved by stacking processed images as additional reference images.

\section{Conclusion}

A novel CNN architecture was presented in this study, along with its unsupervised process that enables image segmentation in an unsupervised manner.
The proposed CNN architecture consists of convolutional filters for feature extraction and differentiable processes for feature clustering, which enables end-to-end network training.
The proposed CNN jointly assigned cluster labels to image pixels and updated the convolutional filters to achieve better separation of clusters using the backpropagation of the proposed loss to the normalized responses of convolutional layers.
Furthermore, two applications based on the proposed segmentation method were introduced: segmentation with scribbles as user input and utilization of reference images.
The experimental results on the PASCAL VOC 2012 segmentation benchmark dataset \cite{everingham2015pascal} and BSD500 \cite{amfm_pami2011} demonstrated the effectiveness of the proposed method for completely unsupervised segmentation.
The proposed method outperformed classical methods for unsupervised image segmentation such as $k$-means clustering and a graph-based segmentation method, which verfied the importance of feature learning.
Furthermore, the effectiveness of the proposed method for image segmentation with user input and utilization of reference images was validated by additional experimental results on the PASCAL VOC 2012, BSD500, and iCoseg \cite{batra2010icoseg} datasets.
A potential application of the proposed method to an efficient video segmentation system was demonstrated.

\ifCLASSOPTIONcaptionsoff
  \newpage
\fi

\bibliographystyle{ieeetr}
\bibliography{unsupervised_segmentation}
\end{document}